\documentclass[11pt]{article}

\usepackage[final]{acl}
\usepackage{tabularx}
\usepackage{times}
\usepackage{latexsym}

\usepackage[T1]{fontenc}

\usepackage[utf8]{inputenc}

\usepackage{microtype}
\usepackage{amsmath}

\usepackage{inconsolata}

\usepackage{graphicx}
\usepackage{booktabs}
\usepackage{multirow}
\usepackage{tabularx}
\usepackage{amssymb}
\usepackage[capitalise,noabbrev,nameinlink]{cleveref}
\usepackage{subfig}
\graphicspath{{../images/}{images/}}

\title{Less Is Moral: A CHARMing Framework for Moral Foundations Detection in Endorsement Behaviour}

\author{
 \textbf{Huixiang Fu},
 \textbf{Marian-Andrei Rizoiu}
\\
Behavioral Data Science, University of Technology Sydney
\\
 \small{
   \textbf{Correspondence:} \href{mailto:huixiang.fu@student.uts.edu.au}{\texttt{huixiang.fu@student.uts.edu.au}}
 }
}

\begin{document}
\maketitle

\begin{abstract}
Moral language plays a central role in shaping online endorsement and the diffusion of information, yet existing moral foundation detection systems often suffer from poor cross-domain generalization, weak rationale grounding, and reliance on costly prompting-based large language models (LLMs).
We introduce CHARM, a MA\textbf{C}- and \textbf{H}ate-speech-\textbf{A}ware \textbf{R}ationale-aligned \textbf{M}oral foundation detection framework built on a lightweight fine-tuned LLM, which integrates complementary moral grounding, rationale alignment, and polarity-aware hate speech signals to support more robust and faithful moral prediction.
Unlike prior dictionary-, fine-tune-, or prompt-based detectors, which decouple computation from psychological theory, CHARM is built so that each component --- MAC cross-attention, rationale alignment, and hate-speech modulation --- operationalizes a distinct psychological construct.
Using a 30\% subsample of the MFTC, MFRC, and News training pools together with the richer supervision in MFTCXplain, CHARM improves AUC by up to 15.3\% in-domain, surpasses the supervised baselines on every out-of-domain dataset in both AUC and F1, and offers a scalable, low-cost alternative to prompting-based LLM detectors.
We further apply CHARM to large-scale COVID-19 discourse on Twitter and show that moral value alignment is strongly associated with online endorsement behavior. 
By making moral framing measurable at scale, CHARM offers a practical tool for studying the spread of morally charged misinformation\footnote{Code and additional materials: \url{https://github.com/HuixiangF/CHARM/}}.
\end{abstract} 

\section{Introduction}
\label{introduction}
In March 2021, the \citet{ccdh2021dozen} named a ``Disinformation Dozen'' responsible for up to 65\% of anti-vaccine content on Facebook and Twitter --- including Joseph Mercola, whose posts claiming hydrogen peroxide could treat COVID-19 were shared over 4{,}600 times on Facebook alone.
\emph{Why does such content attract so much endorsement?}
Prior work links sharing to alignment between a post's moral framing and the audience's values \citep{Abdurahman2025TargetingAM}; 
\citet{RAO2025} further show that during COVID-19, pseudo-experts (the Disinformation Dozen among them) used markedly more negative moral framing than public health experts --- especially along the care/harm and authority/subversion dimensions of Moral Foundations Theory (MFT) \citep{haidt2001emotional,graham2013moral}.
Two questions remain open:
\textbf{(i)~is endorsement driven by the moral framing of content creators}, and
\textbf{(ii)~does the resulting endorsement network exhibit moral homophily between users and the creators they endorse?}

Answering either at population scale requires a reliable way to detect moral framing in text. Computational moral foundation detection has evolved from lexicon-based methods \citep{graham2012mfd,hopp2021extended} to supervised fine-tuning with pretrained language models \citep{nguyen2024mformer,preniqi2024moralbert}, and more recently to prompting LLMs for moral classification \citep{chen2025mova,skorski2025moral}.

Three gaps persist.
(1)~\textbf{Detection decoupled from theory.} Moral cognition is grounded in two complementary psychological frameworks --- MFT and Morality-as-Cooperation (MAC)  \citep{curry2019mapping,curry2019cooperate,graham2011mapping,haidt2012righteous} --- yet fine-tuned detectors inherit only the MFT label vocabulary, collapsing this richer structure into flat classification \citep{araque2022libertymfd,reinig2024survey}. LLM-based detectors substitute statistical pretraining patterns for human moral deliberation \citep{chen2025mova,hendrycks2021aligning,jiang2025delph,amirizaniani2024norm,lyu2024faithful}, producing labels with little auditable evidence.
(2)~\textbf{Polarity and antisocial coupling ignored.} Negative moral framing elevates endorsement \citep{RAO2025}, and moral language is tightly coupled to antisocial expression online \citep{kennedy2023moral,brady2019ideological} --- yet most detectors collapse virtue and vice into a single foundation label and treat hate-speech detection as an unrelated task.
(3)~\textbf{Robustness--efficiency tension.} Fine-tuned models achieve strong in-domain accuracy but generalize poorly; prompt-based LLMs are more robust but expensive and depend on closed APIs.

We address these gaps with \textbf{CHARM} (MA\textbf{C}- and \textbf{H}ate-speech-\textbf{A}ware \textbf{R}ationale-aligned \textbf{M}oral foundation): a two-stage framework on a LoRA-adapted LLaMA-3.1-8B backbone in which each architectural component operationalizes a distinct psychological construct --- MAC cross-attention introduces cooperation-domain structure as a complementary inductive bias, rationale-aligned pooling grounds prediction in human moral-deliberation traces, and FiLM hate-speech modulation captures the documented moral--antisocial coupling. 
This work makes three contributions:

(1) \textbf{Theory-integrated architecture.} To our knowledge, CHARM is the first moral foundation detector to embed psychological structure in the model architecture rather than only the label space, departing from the dictionary-, fine-tune-, and prompt-based traditions that treat moral detection as classification disconnected from the theories whose labels they reuse. Ablations identify MAC grounding as the strongest cross-domain inductive bias, with the largest OOD F1 drops when removed (e.g., ARG .54$\rightarrow$.43, VIG .67$\rightarrow$.50).

(2) \textbf{Efficient, polarity-aware, and faithful.} CHARM is open-weight and uses a 30\% subsample of the MFTC, MFRC, and News training pools, together with MFTCXplain, which provides polarity, rationale, and hate-speech supervision. This complete system improves AUC by up to 29.5\% in-domain and 5.1\% out-of-domain, surpasses the supervised baselines on every OOD dataset in both AUC and F1, and remains competitive with prompting-based LLMs at a fraction of the per-inference cost. Its 10-dimensional polarity-aware head improves over the strongest polarity-aware baseline by .33 AUC. Token-level rationale spans align closely with human annotations and causally support predictions, supplying auditable evidence behind each decision.

(3) \textbf{Moral framing predicts endorsement and network homophily at scale.} On COVID-19 Twitter discourse, CHARM shows that producer-side moral framing predicts endorsement beyond behavioural and network features (author-only moral AUC .84 vs.\ retweeter-only .72), and that the network is morally assortative across all foundations, most strongly along loyalty ($r=.46$).

\section{Related Work}

\subsection{Moral Foundations Theory (MFT) and Morality-as-Cooperation (MAC)}
MFT organizes human moral reasoning around five evolved foundations \citep{haidt2004intuitive,graham2009liberals,graham2013moral}: \textit{care/harm}, \textit{fairness/cheating}, \textit{loyalty/betrayal}, \textit{authority/subversion}, and \textit{purity/degradation} --- the first two termed \textit{individualizing} and the latter three \textit{binding} foundations --- with a later extension adding \textit{liberty/oppression} \citep{iyer2012understanding}.

MAC \citep{curry2016morality} grounds morality in evolutionary game theory, identifying seven cooperation-based domains: \textit{family values} (kin altruism), \textit{group loyalty} (coalition support), \textit{reciprocity} (mutual exchange), \textit{heroism} (costly altruism), \textit{deference} (hierarchy), \textit{fairness} (equitable allocation), and \textit{property rights} (ownership). Although developed through distinct methodologies, the two frameworks are largely complementary: several MAC domains map onto MFT foundations (e.g., group loyalty onto loyalty/betrayal, deference onto authority/subversion), while MAC additionally captures cooperative moral types absent from MFT, such as kin altruism, heroism, and property rights \citep{curry2019mapping}.

\subsection{Moral Foundation Detection}
\label{sec:mft_detection}

Computational moral foundation detection spans three families. \textit{Lexicon-based methods} rely on predefined moral dictionaries --- MFD \citep{graham2012mfd}, MFD~2.0  \citep{frimer2019moral}, and eMFD  \citep{hopp2021extended} --- and scale well but struggle with contextual and implicit expressions. \textit{Fine-tuned models} such as MoralBERT   \citep{preniqi2024moralbert} and MFormer \citep{nguyen2024mformer} improve contextual sensitivity but generalize poorly across domains; recent work addresses this through auxiliary supervision and domain adaptation, e.g., DAMF \citep{guo2023damf} and ME$^2$-BERT \citep{zangari2025me2bert}. \textit{Prompting-based LLMs} such as MoVa \citep{chen2025mova} improve transferability but depend on closed, proprietary APIs. Across these families, moral detection remains a label-prediction task with limited theoretical grounding and auditable evidence.
Recent work addresses the latter gap with rationale-rich hate-speech datasets. MFTCXplain provides a multilingual benchmark with multi-hop explanations for LLM moral reasoning~\citep{trager2025mftcxplain}, while \citet{vargas2026selfexplaining} provide Brazilian Portuguese moral rationales for self-explaining hate-speech detection. CHARM uses these resources for rationale-rich training and cross-lingual evaluation, respectively, and adds MAC grounding for cross-domain MFT detection.

\section{CHARM Framework}
\begin{figure*}[t]
  \centering
  \includegraphics[width=\textwidth]{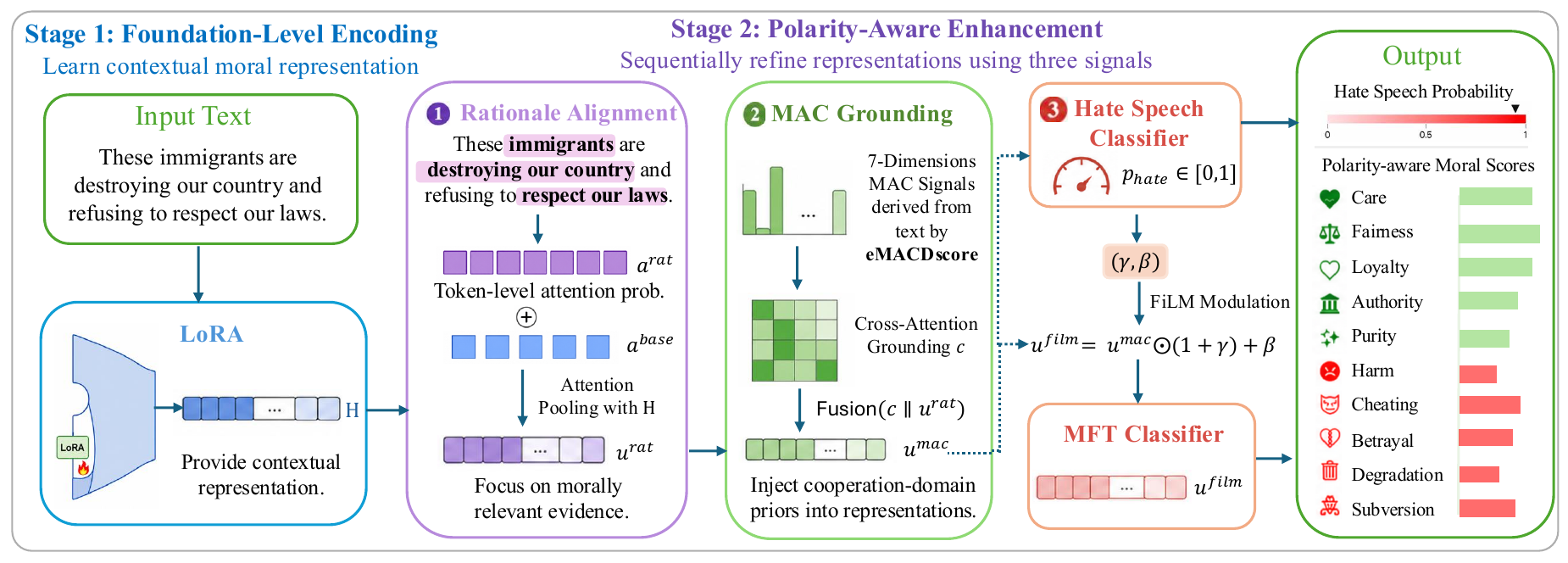}
  \caption{Overview of CHARM, a two-stage framework for polarity-aware moral foundation detection. Stage 1 trains a LoRA adapter to learn foundation-level moral semantics. Stage 2 refines the text representation using rationale alignment, MAC-guided cross-attention, and hate-speech-aware FiLM modulation. The model jointly predicts 10 polarity-aware moral dimensions and hate speech.}
  \label{fig:architecture}
\end{figure*}

Each component of CHARM operationalizes a distinct psychological construct: (i) the MFT foundation classifier targets the standard moral foundation taxonomy; (ii) the MAC cross-attention module introduces cooperation-domain structure as a complementary inductive bias; (iii) the rationale-alignment module grounds prediction in human moral-deliberation traces; and (iv) the hate-speech modulation captures the documented coupling between moral language and antisocial expression.
As shown in \cref{fig:architecture}, \textsc{CHARM} consists of two stages: (i) a LoRA-based encoder adaptation stage for learning foundation-level moral representations, and (ii) a polarity-aware classification stage that integrates rationale supervision, MAC grounding, and hate-speech modulation.

\subsection{Problem Statement}

Given a dataset
$\mathcal{D} = \{(x_i, y_i^{(m)}, y_i^{(h)}, r_i, g_i)\}_{i=1}^{N}$,
where $y_i^{(m)} \in \{0,1\}^{10}$ denotes polarity-aware moral labels, $y_i^{(h)} \in \{0,1\}$ the hate-speech label, $r_i$ the rationale mask, and $g_i \in \mathbb{R}^{7}$ the MAC vector, the goal is to learn a polarity-aware moral classifier.

\subsection{Foundation-Level Moral Encoding}

Moral datasets differ substantially in annotation granularity: most provide only coarse-grained foundation labels, while a smaller subset additionally includes polarity annotations. 
To leverage all available supervision consistently, we adopt a two-stage training strategy. 
Stage~1 learns transferable foundation-level moral representations using supervision from the five MFT foundations, while Stage~2 extends the model with polarity-aware signals built on top of the frozen Stage~1 encoder.
The encoder is fine-tuned using LoRA~\citep{hu2022lora}. A lightweight classifier predicts five foundation logits $\hat{y} \in \mathbb{R}^{5}$, optimized with: 
\[
\mathcal{L}_{\mathrm{moral}}^{(5d)}
=
\frac{1}{5}
\sum_{k=1}^{5}
\mathrm{BCE}(\hat{y}_k,\; y_k^{(5)}).
\]
\subsection{Polarity-Aware Moral Classification}
Stage~2 builds on the frozen LoRA-adapted encoder from Stage~1 and performs polarity-aware moral classification using rationale alignment, MAC grounding, and hate-speech modulation.
\subsubsection{Human Rationale Supervision}

\paragraph{Rationale Selector.}
Given an input $x$, the LoRA encoder from Stage~1 produces text representations $H = \mathrm{Enc}(x) \in \mathbb{R}^{L \times d}$, where $L$ is the sequence length and $d$ is the hidden dimension. 
A rationale selector predicts token-level rationale logit $z_t$ and probability $p_t$, indicating whether the token belongs to a human-annotated rationale span. 
The selector is pretrained using token-level rationale supervision with binary cross-entropy:
\[
\mathcal{L}_{\mathrm{rat}}
=
\frac{1}{|\mathcal{V}|}
\sum_{t \in \mathcal{V}}
\mathrm{BCE}(\sigma(z_t), r_t),
\]
where $r_t \in \{0,1\}$ denotes the aligned rationale label and $\mathcal{V}$ represents non-padding positions. To encourage contiguous rationale predictions, we additionally apply a total variation regularizer:
\[
\mathcal{L}_{\mathrm{tv}}
=
\frac{1}{|\mathcal{A}|}
\sum_{t \in \mathcal{A}}
(p_t - p_{t-1})^2,
\]
where $\mathcal{A}$ denotes adjacent valid token positions.

\paragraph{Rationale-Steered Attention Pooling.}
We combine the encoder attention distribution 
$a^{\mathrm{base}}$ with the predicted rationale 
distribution $a^{\mathrm{rat}}$ through a learnable 
interpolation $
a = (1 - \alpha)a^{\mathrm{base}}
+ \alpha a^{\mathrm{rat}},
$
where $\alpha \in [0,1]$ is a learnable interpolation coefficient parameterized through a sigmoid function. The final text representation is obtained by weighted pooling over token representations, where 
$u^{\mathrm{rat}} = \sum_{t=1}^{L} a_t h_t.$ 
This encourages the model to focus on human-identified moral evidence while retaining contextual information from the encoder representations.
\subsubsection{MAC-Theory Grounding}
We introduce the Moral-As-Cooperation (MAC) grounding as an auxiliary supervision signal. 
Each sample is associated with seven MAC cooperation dimensions from eMACDscore~\citep{malik2025emacd}, a lightweight dictionary-based moral mining tool that scores text over the seven MAC cooperation domains (see \cref{app:emacd} for details). Each dimension provides a probability score $\hat{p}_i$ and a sentiment-polarity score $\hat{s}_i$ that capture the strength and direction of the cooperative signal. After normalization, the two scores are multiplied as $g_i = \tilde{p}_i \cdot \tilde{s}_i$, forming a MAC vector $g \in \mathbb{R}^{7}$. Each dimension of $g$ is then projected into the hidden space to produce seven MAC tokens $M \in \mathbb{R}^{7 \times d}$.
The MFT labels are represented as ten learnable queries $Q \in \mathbb{R}^{10 \times d}$, corresponding to the virtue and vice poles of the five foundations. Cross-attention between $Q$ and $M$ produces ten moral representations:
\[
C = \mathrm{softmax}\!\left(\frac{QM^{\top}}{\sqrt{d}}\right) M \in \mathbb{R}^{10 \times d},
\]
along with an attention map $W = \mathrm{softmax}(QM^{\top}/\sqrt{d}) \in \mathbb{R}^{10 \times 7},$
where $W_{ij}$ denotes how strongly the $i$-th moral dimension attends to the $j$-th MAC dimension. 
The representations in $C$ are then pooled into a single MAC vector $c$ using learnable attention-based weighting. Specifically, each moral representation $C_i$ is assigned a scalar attention score:
\[
a_i
=
\frac{
\exp(w^\top C_i)
}{
\sum_{k=1}^{10}
\exp(w^\top C_k)
},
\]
where $w \in \mathbb{R}^{d}$ is a learnable parameter vector. The final pooled MAC representation is then computed as
$c
=
\sum_{i=1}^{10}
a_i\, C_i
\in \mathbb{R}^{d}.$
Finally, the MAC vector $c$ is fused with the rationale-guided text representation $u^{\mathrm{rat}}$ to obtain the MAC-enhanced moral representation $u^{\mathrm{mac}}$:
\[
u^{\mathrm{mac}} = \mathrm{FusionMLP}([u^{\mathrm{rat}} \,\|\, c]) \in \mathbb{R}^{d}.
\]
This formulation injects cooperation-domain structure into the moral representation, enabling the model to capture complementary interactions between MAC signals and moral foundations.
\subsubsection{Hate Speech Modulation}

Prior work has shown that moral language and hate speech are closely related, with certain moral dimensions disproportionately associated with hateful rhetoric~\citep{vargas2026selfexplaining}. Motivated by this connection, we introduce hate speech as an auxiliary modulation signal for polarity-aware moral prediction.
Given the MAC-grounded representation $u^{\mathrm{mac}} \in \mathbb{R}^{d}$, a hate-speech classifier first predicts a hate probability $\hat{y}^{(h)} = \sigma(w_h^{\top} u^{\mathrm{mac}})$. The hate speech head is trained using binary cross-entropy loss.
Following Feature-wise Linear Modulation (FiLM)~\citep{perez2018film}, the predicted hate probability is transformed into channel-wise scaling and shifting parameters:
\[
\gamma = \tanh(f_{\gamma}(\hat{y}^{(h)})),
\qquad
\beta = f_{\beta}(\hat{y}^{(h)}),
\]
where $\gamma \in [-1,1]^d$ and $\beta \in \mathbb{R}^{d}$.
We then modulate $u^{\mathrm{mac}}$ through $u^{\mathrm{film}} = u^{\mathrm{mac}} \odot (1+\gamma) + \beta$, allowing hate speech information to act as a soft conditioning signal over the moral representation.
The final polarity-aware moral prediction is computed as a 10-dimensional probability vector, $\hat{y}^{(m)} = \sigma(W_m u^{\mathrm{film}}) \in (0,1)^{10}$. The corresponding classification objective is:
\[
\mathcal{L}_{\mathrm{moral}}^{(10d)}
=
\frac{1}{10}
\sum_{j=1}^{10}
\mathrm{BCE}(\hat{y}^{(m)}_j,\; y^{(m)}_j).
\]




\section{Experiments}
\label{sec:experiments}


\begin{table*}[t]
\centering
\small
\begin{tabularx}{\textwidth}{@{}>{\centering\arraybackslash}p{1.4cm} l r >{\hsize=0.85\hsize}X >{\hsize=1.15\hsize}X@{}}
\toprule
\textbf{Split} & \textbf{Dataset} & \textbf{\# Inst.} & \textbf{Text Type} & \textbf{Content Focus} \\
\midrule
\multirow{4}{*}{{\textit{In-domain}}}
& MFTC~\citep{hoover2020moral} & 34,987 & Twitter posts & Socio-political discourse  \\
& MFRC~\citep{trager2022moral} & 17,886 & Reddit comments & Politics and everyday morality \\
& News~\citep{hopp2021extended} & 34,262 & News articles & Diverse news topics from GDELT \\
& \textsuperscript{$\dagger$}MFTCXplain~\citep{trager2025mftcxplain} & 3,245 & Multilingual tweets & Moral reasoning over hate speech \\
\midrule
\multirow{5}{*}{{\textit{OOD}}}
& \textsc{SC}~\citep{forbes2020social} & 29,239 & Rules-of-thumb & Everyday social and moral norms \\
& \textsc{VIG}~\citep{clifford2015moral} & 132 & Psychology vignettes & Foundation-violating scenarios \\
& \textsc{ARG}~\citep{kobbe2020exploring} & 320 & Debate arguments & Pro/con stances on 16 topics \\
& \textsc{MIC}~\citep{ziems2022moral} & 11,375 & Chatbot RoTs & Human--chatbot moral biases \\
& \textsuperscript{$\ddagger$}\textsc{HateBR}~\citep{vargas2026selfexplaining} & 5,040 & Instagram comments & Brazilian political hate speech \\
\bottomrule
\end{tabularx}
\caption{Full profile of datasets used for training and evaluation. \# Inst is the original dataset size.
\textsuperscript{$\dagger$}MFTCXplain contains four languages (Portuguese, Persian, Italian, English).
\textsuperscript{$\ddagger$}HateBR is in Brazilian Portuguese.}
\label{tab:data-profile-full}
\end{table*}

\paragraph{Datasets.}
We use nine datasets covering diverse forms of morally charged discourse. Dataset statistics and label distributions are provided in \cref{tab:data-profile-full} and \cref{app:label-distributions}. For all datasets, we precompute 7-dimensional MAC signals using eMACDscore.
Training proceeds in two stages. Stage~1 jointly trains on four in-domain datasets using foundation-level supervision, while Stage~2 further fine-tunes on MFTCXplain with polarity, rationale, and hate-speech annotations. During Stage~1, polarity labels in MFTCXplain are collapsed into foundation categories, while the same data split is retained across both stages to prevent leakage.
Following \textsc{Mformer}, we retain the original train/test splits and use 30\% of the training data for the remaining in-domain datasets. We exclude liberty/oppression due to inconsistent supervision across datasets. The five out-of-domain datasets are reserved for evaluation under a shared validation setting.

\paragraph{Backbone selection.}
We compare \textsc{LLaMA-3.1-8B}\footnote{\url{https://huggingface.co/meta-llama/Llama-3.1-8B}} and \textsc{Qwen3-8B}\footnote{\url{https://huggingface.co/Qwen/Qwen3-8B}}, and find that \textsc{LLaMA-3.1-8B} consistently performs better across settings. We therefore use it as the backbone in all main experiments. Additional experimental details and \textsc{Qwen3-8B} results are provided in \cref{Experimental-Setup} and \cref{app:qwen-results}, respectively.

\paragraph{Baselines.}
We compare CHARM with five representative baselines spanning fine-tuned and zero-shot approaches. 
Because these models differ in their training objectives and available supervision, 
\cref{tab:training-corpora} explicitly summarizes their training paradigms and training corpora to contextualize the comparison.

\begin{table*}[t]
\centering
\small
\renewcommand{\arraystretch}{1.15}
\setlength{\tabcolsep}{4pt}

\begin{tabularx}{\textwidth}{
>{\raggedright\arraybackslash}p{0.20\textwidth}
>{\raggedright\arraybackslash}p{0.10\textwidth}
>{\raggedright\arraybackslash}p{0.28\textwidth}
>{\raggedright\arraybackslash}X}
\toprule
\textbf{Model} & \textbf{Training} & \textbf{Training data} & \textbf{Description} \\
\midrule

\textsc{MoralBERT}\newline\cite{preniqi2024moralbert}
& Fine-tuned
& MFTC
& BERT-based model for polarity-aware moral foundation classification. \\

\textsc{Mformer}\newline\cite{nguyen2024mformer}
& Fine-tuned
& MFTC, MFRC, and News
& RoBERTa-based classifiers trained for multi-domain moral foundation detection. \\

\textsc{MoVa}\newline\cite{chen2025mova}
& Zero-shot
& None
& LLM prompting framework for joint moral foundation prediction. \\

Qwen-32B-Instruct\footnotemark
& Zero-shot
& None
& Large instruction-tuned LLM evaluated through direct zero-shot prediction. \\

Tuning-GPT4o-mini\newline\cite{chen2025mova}
& Fine-tuned
& 30\% of MFTC, MFRC, and News
& Instruction-tuned LLM fine-tuned for moral foundation prediction. \\

\textsc{CHARM}
& Fine-tuned
& MFTCXplain and 30\% of MFTC, MFRC, and News
& Theory-grounded model with rationale and polarity-aware supervision. \\

\bottomrule
\end{tabularx}

\caption{Overview of the compared models and their training configurations. 
The models differ in training paradigm and available training corpora; these differences are made explicit to contextualize the performance comparison.}
\label{tab:training-corpora}
\end{table*}

\footnotetext{\url{https://huggingface.co/Qwen/Qwen2.5-32B-Instruct}}

\paragraph{Metrics.}
We evaluate the model along two axes: classification performance and rationale quality. For classification, we report F1 and AUC on both moral foundation prediction and hate speech detection. For rationale quality, we follow the ERASER protocol~\cite{deyoung2020eraser} to assess \emph{plausibility} and \emph{faithfulness}. Definitions of all evaluation metrics are provided in \cref{app:metric-definitions}.

\subsection{Moral Detection Performance}
\label{sec:main-results}

\begin{table*}[t]
  \centering
  \small
  \renewcommand{\arraystretch}{1.0}
  \setlength{\tabcolsep}{4pt}
  \begin{tabular}{llcccc|ccccc}
    \toprule
    & & \textbf{MFTC} & \textbf{MFRC} & \textbf{News} & \textbf{MFTCXpl.} & \textbf{ARG} & \textbf{SC} & \textbf{MIC} & \textbf{VIG} & \textbf{HateBR} \\
    \midrule
    \multirow{6}{*}{\rotatebox{90}{\scriptsize AUC}}
    & \textsc{CHARM}              & .87 & \textbf{.91} & \textbf{.83} & \textbf{.79} & \textbf{.87} & \textbf{.77} & .73 & .89 & \textbf{.82} \\
    & \textsc{MoVa}$^\dagger$      & .77 & .80 & .68 & .75 & \textbf{.87} & .76 & \textbf{.75} & \textbf{.96} & .78 \\
    & Qwen-32B-Inst.               & .64 & .63 & .55 & .62 & .72 & .66 & .63 & .88 & .63 \\
    & \textsc{Mformer}$^\dagger$   & \textbf{.89} & .84 & .72 & .61 & .86 & .70 & .70 & .81 & .62 \\
    & \textsc{MoralBERT}$^\dagger$ & .75 & .79 & .61 & .60 & .77 & .70 & .69 & .80 & .54 \\
    & Tuning-GPT4o-mini$^\dagger$  & --  & --  & --  & --  & .85 & .73 & .73 & .92 & -- \\
    \midrule
    \multirow{5}{*}{\rotatebox{90}{\scriptsize F1}}
    & \textsc{CHARM}     & .72 & \textbf{.72} & \textbf{.55} & \textbf{.59} & \textbf{.54} & .46 & .50 & .67 & \textbf{.38} \\
    & \textsc{MoVa}      & .57 & .48 & .37 & .52 & .49 & \textbf{.52} & \textbf{.51} & .69 & .36 \\
    & Qwen-32B-Inst.     & .47 & .40 & .26 & .41 & .52 & .42 & .40 & \textbf{.79} & .30 \\
    & \textsc{Mformer}   & \textbf{.77} & .66 & .53 & .50 & .51 & .43 & .44 & .52 & .31 \\
    & \textsc{MoralBERT} & .56 & .49 & .33 & .38 & .47 & .37 & .41 & .32 & .24 \\
    \bottomrule
  \end{tabular}
  \caption{
    F1 and AUC across nine datasets. Best results per dataset are shown in \textbf{bold}. Baseline AUC values ($^\dagger$) are taken from \citet{chen2025mova}; F1 scores are independently reproduced by running each model. ``--'' indicates results not reported and not reproducible. Per-foundation breakdowns are provided in \cref{app:per-label}. Qwen-32B-Instruct and \textsc{MoVa} are evaluated using zero-shot prompting, whereas the other models use task-specific fine-tuning.
  }
  \label{tab:main-results}
\end{table*}

\paragraph{In-domain performance.}
As shown in \cref{tab:main-results}, CHARM performs strongly across the in-domain benchmarks, achieving the best AUC and F1 on MFRC, News, and MFTCXplain.
On the three corpora shared with \textsc{Mformer}---MFTC, MFRC, and News---CHARM uses only 30\% of the available training instances, supplemented by the richer supervision from MFTCXplain.
CHARM improves over \textsc{Mformer} on MFRC and News by .07 and .11 AUC, respectively, with corresponding gains in F1. MFTC is the main exception: \textsc{Mformer} achieves a higher F1 (.77 vs.\ .72) and slightly higher AUC (.89 vs.\ .87), with the F1 difference primarily concentrated in the Authority foundation
(see \cref{app:per-label}).
Overall, CHARM's gains are strongest on MFRC and News, while its performance on MFTC remains competitive rather than uniformly superior.

\paragraph{Generalization under distribution shift.}
CHARM also generalizes robustly to unseen domains, although its advantage varies across baseline families.
Among fine-tuned baselines, CHARM consistently outperforms \textsc{Mformer} across all five OOD datasets in both AUC and F1.
Compared with Tuning-GPT4o-mini, CHARM matches or exceeds the reported AUC on four of the five OOD datasets, with VIG as the exception.
Together, these results indicate that CHARM maintains strong cross-domain performance despite its limited supervision on MFTC, MFRC, and News.
The comparison with zero-shot LLMs is more nuanced. Against \textsc{MoVa}, CHARM performs better on ARG (F1: .54 vs.\ .49; equal AUC), remains competitive on SC and HateBR, but trails on MIC and VIG.
Similarly, CHARM outperforms Qwen-32B-Instruct in AUC across all nine datasets and in F1 on eight of nine, while Qwen shows a clear advantage on VIG (.79 vs.\ .67 F1).
These results suggest that CHARM provides more consistent performance across domains, whereas zero-shot LLMs can be stronger on particular datasets.

One possible explanation lies in input characteristics and inference strategy. SC and MIC contain relatively short inputs (14 and 17 tokens on average), whereas ARG contains substantially longer inputs (69.6 tokens on average), where moral evidence may be distributed across the context.
Task-specific fine-tuning and rationale supervision may therefore be particularly useful for capturing dispersed moral cues, while prompting-based models remain competitive on shorter inputs.
Consistent with this interpretation, CHARM also outperforms \textsc{MoVa} on the Care foundation across SC and MIC, where its hate-speech-aware supervision may improve sensitivity to harm-related cues.

\begin{figure}[t]
  \centering
  \includegraphics[width=1.00\columnwidth]{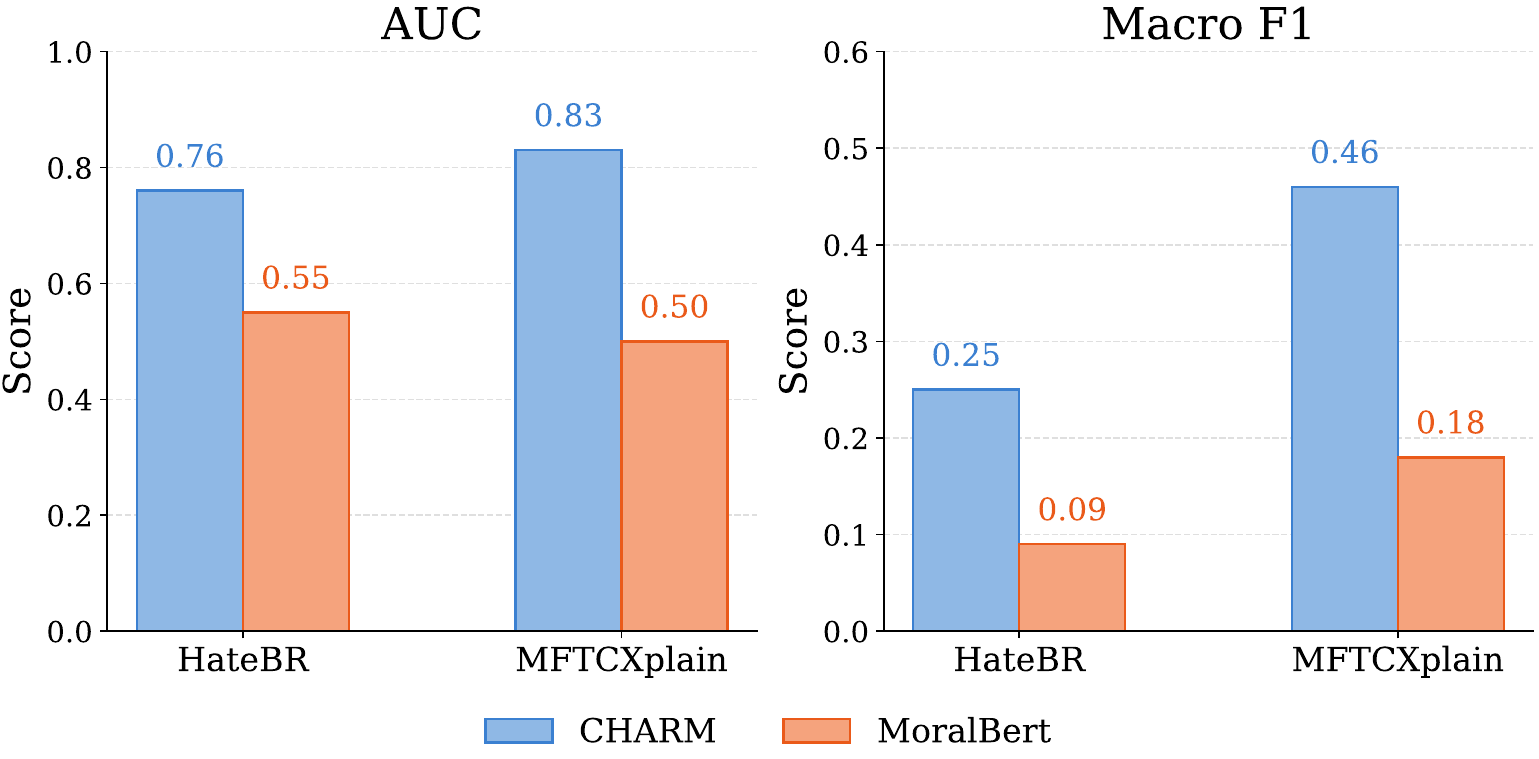}
  \caption{AUC and Macro-F1 comparison between CHARM and \textsc{MoralBERT} for 10-dimension moral polarity classification on MFTCXplain and HateBR (only datasets with 10-dimensional moral annotations).}
  \label{fig:ten-dim-results}
\end{figure}

\paragraph{Moral polarity classification.}
As shown in \cref{fig:ten-dim-results}, CHARM clearly outperforms \textsc{MoralBert} on both datasets, achieving AUC improvements of .33 on MFTCXplain and .21 on HateBR, alongside even larger gains in F1. Notably, CHARM also performs well on the Portuguese HateBR dataset, even though it was not designed for multilingual learning. This suggests that training on the multilingual MFTCXplain dataset may provide an unintended cross-lingual benefit.

\paragraph{Inference cost.}
Beyond predictive performance, CHARM and prompting-based approaches differ in deployment cost.
\textsc{MoVa} incurs inference-time API costs of approximately \$.066 and \$.065 per 1K input tokens on SC and MIC, respectively, with costs incurred for each inference pass.
By contrast, CHARM is fine-tuned once on an open-source backbone and requires no per-example API calls at inference time.

\subsection{Ablation Studies}
\label{sec:ablation}
We ablate MAC grounding, rationale supervision, and hate-speech supervision; F1 results are shown in \cref{tab:ablation}, with AUC results in \cref{app:ablation-auc}. We also assess the contributions of CHARM's training corpora through a leave-one-corpus-out analysis, with full results in \cref{app:corpus-ablation}.


\begin{table*}[t]
  \centering
  \small
  \renewcommand{\arraystretch}{1.0}
  \setlength{\tabcolsep}{4pt}
  \begin{tabular}{lccccccccc}
    \toprule
    \textbf{Variant} & \textbf{MFTCXpl.$_{10d}$} & \textbf{MFTCXpl.$_{5d}$} & \textbf{MFTC} & \textbf{MFRC} & \textbf{News} & \textbf{ARG} & \textbf{SC} & \textbf{MIC} & \textbf{VIG} \\
    \midrule
    baseline         & .41 & .56 & .68 & .56 & .50 & .45 & .30 & .43 & .47 \\
    w/o hate speech  & .45 & .59 & .64 & .55 & .55 & .53 & .44 & .45 & .65 \\
    w/o MAC          & .46 & .59 & .66 & .53 & .47 & .43 & .41 & .44 & .50 \\
    w/o rationale    & .42 & .57 & .68 & .65 & .55 & .53 & .46 & .47 & .60 \\
    \midrule
    \textsc{CHARM}   & \textbf{.47} & \textbf{.59} & \textbf{.72} & \textbf{.69} & \textbf{.55} & \textbf{.54} & \textbf{.46} & \textbf{.50} & \textbf{.67} \\
    \bottomrule
  \end{tabular}
  \caption{Ablation study reporting F1 scores across evaluation datasets. The baseline applies a task-specific classification head on top of the frozen Stage~1 LoRA encoder using only the 10-dimensional MFT supervision. AUC results show consistent trends and are reported in \cref{app:ablation-auc}.}
  \label{tab:ablation}
\end{table*}

\paragraph{MAC provides the strongest structural inductive bias for cross-domain generalization.}
Removing MAC causes the largest overall degradation across OOD evaluation datasets, with especially severe drops on ARG (.54$\rightarrow$.43) and VIG (.67$\rightarrow$.50). 
Notably, compared with the other signals, SC and MIC are more strongly affected by removing MAC (SC: .46$\rightarrow$.41; MIC: .50$\rightarrow$.44). Both datasets primarily involve socially contextualized norm judgments and rule-of-thumb reasoning rather than explicit moral declarations, suggesting that MAC provides useful structural grounding for implicit moral interpretation under distribution shift.

\paragraph{Rationale supervision improves evidence grounding and prediction faithfulness.}

Removing rationale supervision leads to larger performance drops on datasets with relatively longer and more compositional inputs, particularly MFRC (.69$\rightarrow$.65) and VIG (.67$\rightarrow$.60), while having comparatively smaller effects on shorter-text datasets such as SC. Since rationale supervision is provided at the span level, this pattern suggests that rationale alignment is especially beneficial when morally relevant evidence is distributed across contextual spans rather than concentrated in a single local cue.
This effect is further supported by the rationale evaluation results in \cref{tab:rationale_quality}. Compared with the variant without rationale supervision, CHARM consistently improves performance across all rationale evaluation metrics, indicating that rationale alignment enhances both plausibility and faithfulness, while also improving the causal relevance of the identified rationale spans for prediction.
We further examine representative cases with and without rationale supervision, providing qualitative evidence that rationale supervision helps preserve label-relevant evidence and improves foundation-level predictions (see \cref{app:qualitative-rationales}).

\begin{table}[t]
  \centering
  \small
  \renewcommand{\arraystretch}{1.0}
  \setlength{\tabcolsep}{3pt}
  \begin{tabular}{lccccc}
    \toprule
     & \multicolumn{3}{c}{\textbf{Plausibility}} & \multicolumn{2}{c}{\textbf{Faithfulness}} \\
    \cmidrule(lr){2-4}\cmidrule(lr){5-6}
     & \textbf{IoU}$\uparrow$ & \textbf{TF1}$\uparrow$ & \textbf{AUP}$\uparrow$ & \textbf{Suf}$\downarrow$ & \textbf{Cmp}$\uparrow$ \\
    \midrule
    \textsc{CHARM}    & \textbf{.62} & \textbf{.63} & \textbf{.60} & \textbf{.02} & \textbf{.25} \\
    w/o rationale     & .14 & .12 & .42 & .20 & .04 \\
    \bottomrule
  \end{tabular}
  \caption{Rationale quality evaluation on MFTCXplain. Arrows indicate whether higher ($\uparrow$) or lower ($\downarrow$) is better.}
  \label{tab:rationale_quality}
\end{table}


\paragraph{Hate-speech and moral supervision reinforce each other.}
Removing hate-speech supervision consistently reduces performance across multiple datasets, including MFTCXplain$_{10d}$ (.47$\rightarrow$.45), SC (.46$\rightarrow$.44), and VIG (.67$\rightarrow$.65). The degradation on MFTCXplain$_{10d}$ suggesting that hate-speech supervision can help the model better distinguish fine-grained moral polarity.
\cref{tab:hate_speech} further shows that CHARM achieves the strongest hate-speech detection performance, outperforming both prompted GPT-4o mini and a fine-tuned \textsc{LLaMA} model that is trained without moral supervision. These results indicate a mutually beneficial relationship between moral reasoning and harmful-content identification.
Overall, all three supervision signals contribute complementary gains.

\paragraph{Leave-one-corpus-out analysis.}
The full configuration achieves the strongest average performance
(AUC=.83; F1=.57), with MFRC making the largest contribution. MFTC and
News provide complementary domain coverage. Removing MFTCXplain lowers
average performance and produces the clearest declines on MFTCXplain
and HateBR, confirming the value of its polarity, rationale, and
hate-speech supervision. Variation across individual datasets suggests
corpus-specific interactions rather than uniform gains from every
training source. Overall, the results show that CHARM benefits from both
corpus diversity and richer supervision. Complete results are reported in
\cref{tab:corpus-ablation-auc,tab:corpus-ablation-f1} in
\cref{app:corpus-ablation}.

\begin{table}[t]
  \centering
  \small
  \renewcommand{\arraystretch}{1.0}
  \setlength{\tabcolsep}{3pt}
  \begin{tabular}{lccc}
    \toprule
     & \textbf{Bin.\ F1} & \textbf{Mac.\ F1} & \textbf{AUC} \\
    \midrule
    \textsc{CHARM}              & \textbf{.66} & \textbf{.70} & \textbf{.79} \\
    \textsc{Tuning-Llama-3.1-8B} & .61 & .64 & .71 \\
    GPT-4o mini (0-shot)        & .43 & .58 & .61 \\
    GPT-4o mini (4-shot)        & .47 & .58 & .61 \\
    \bottomrule
  \end{tabular}
  \caption{Hate speech detection results on MFTCXplain.}
  \label{tab:hate_speech}
\end{table}

\section{Moral Alignment and Endorsement Behaviour}

\begin{figure}[t]
  \centering
  \includegraphics[width=\columnwidth]{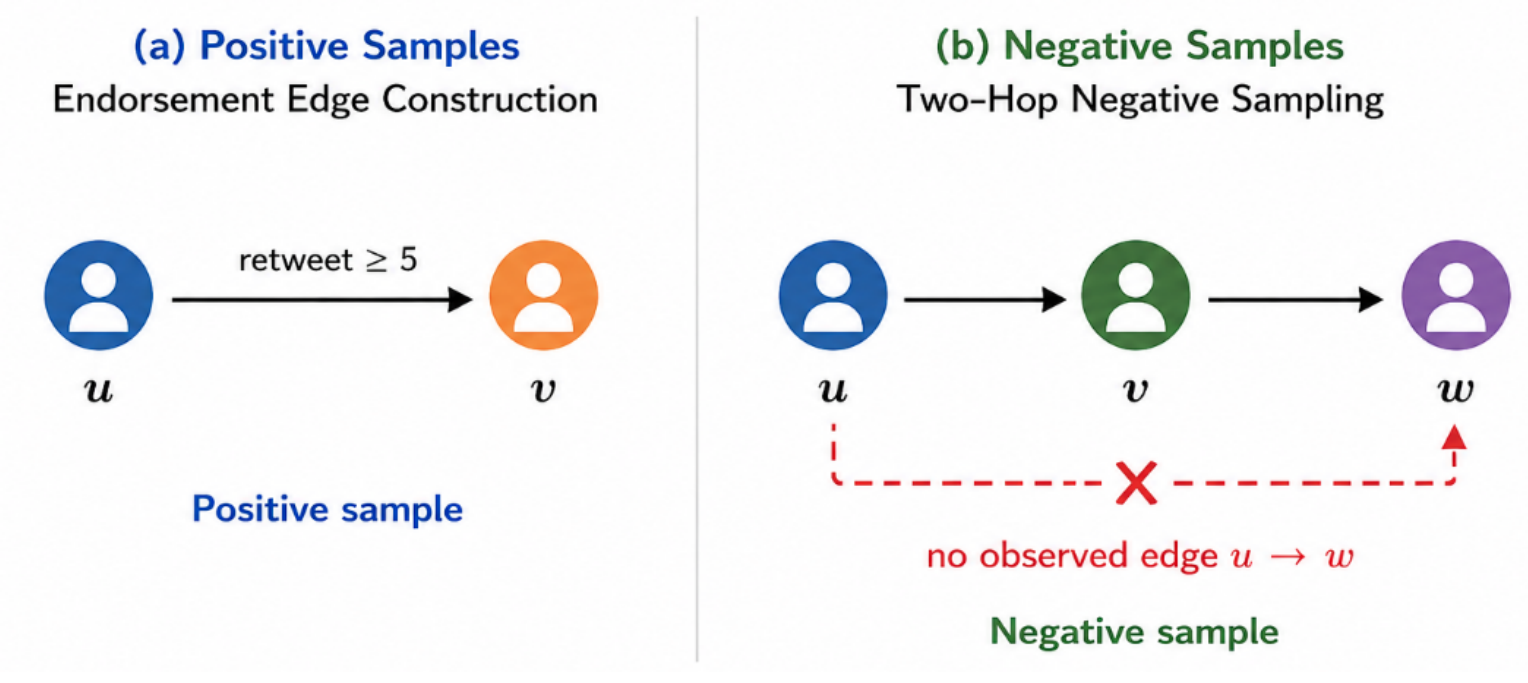}
  \caption{Construction of the endorsement network.}
  \label{fig:endorsement-structure}
\end{figure}

We use CHARM to test, at scale, the two questions raised in \cref{introduction}: \emph{whether producer-side moral framing predicts endorsement, and whether the resulting endorsement network exhibits moral homophily between users and creators}~\citep{dehghani2016purity,vanbavel2021how}. We use a subset of the COVID-19 Twitter dataset introduced by \citet{chen2020tracking}, covering discourse from March 25 to April 16, 2020. Each user's moral profile is constructed by aggregating tweet-level moral scores inferred by \textsc{CHARM}. Although CHARM produces polarity-aware predictions over ten moral dimensions, we collapse scores into five foundation-level representations for endorsement analysis because several polarity dimensions are highly sparse in large-scale Twitter discourse. The full polarity distribution is provided in \cref{app:Endorsement-Network-Construction}. To reduce noise from weak moral expressions, we retain only tweets above the 90th percentile of overall moral intensity.
We then construct a directed endorsement network based on retweet behavior, which is commonly treated as a proxy for endorsement \citep{metaxas2015retweets}. Mentions and quote tweets are excluded due to their ambiguous endorsement semantics. The pair-construction process is illustrated in \cref{fig:endorsement-structure}.

\begin{table}[t]
  \centering
  \small
  \renewcommand{\arraystretch}{1.0}
  \setlength{\tabcolsep}{4pt}
  \begin{tabular}{lccc}
    \toprule
    \textbf{Feature Setting} & \textbf{AUC} & \textbf{Macro F1} & \textbf{Micro F1} \\
    \midrule
    full & \textbf{.92} & \textbf{.84} & \textbf{.84} \\
    w/o moral features & .90 & .82 & .82 \\
    w/o interaction features & .85 & .77 & .77 \\
    author moral only & .84 & .76 & .76 \\
    retweeter moral only & .72 & .66 & .66 \\
    \bottomrule
  \end{tabular}
  \caption{Ablation study for endorsement prediction using interaction, behavioral, and moral-profile features.}
  \label{tab:endorsement-ablation}
\end{table}

\begin{figure*}[t]
  \centering
  \newcommand\myheight{0.125}
  \subfloat[]{\label{fig:shap-full}%
    \includegraphics[height=\myheight\textheight]{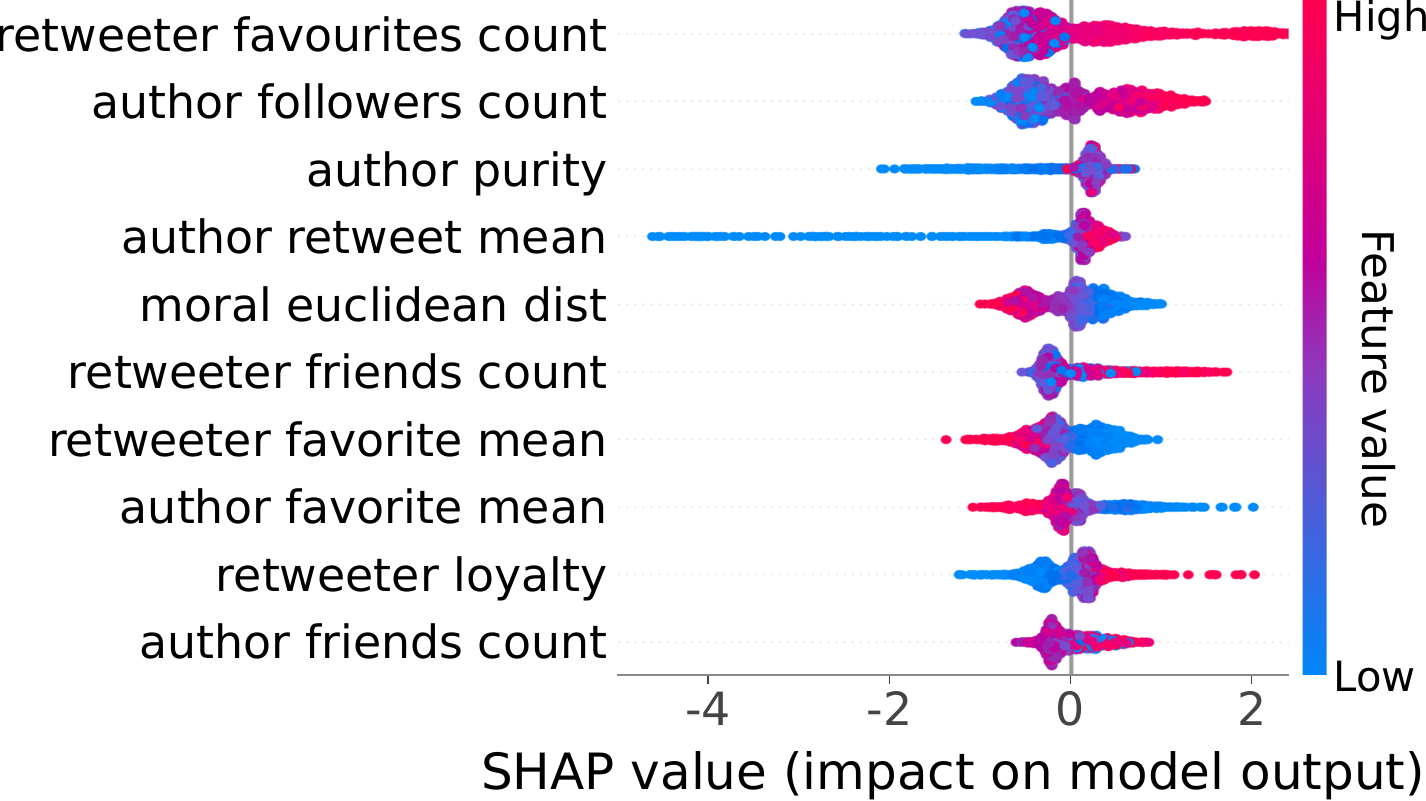}%
    \label{subfig:shap-full}%
  }%
  \hfill
  \subfloat[]{\label{fig:shap-moral}%
    \includegraphics[height=\myheight\textheight]{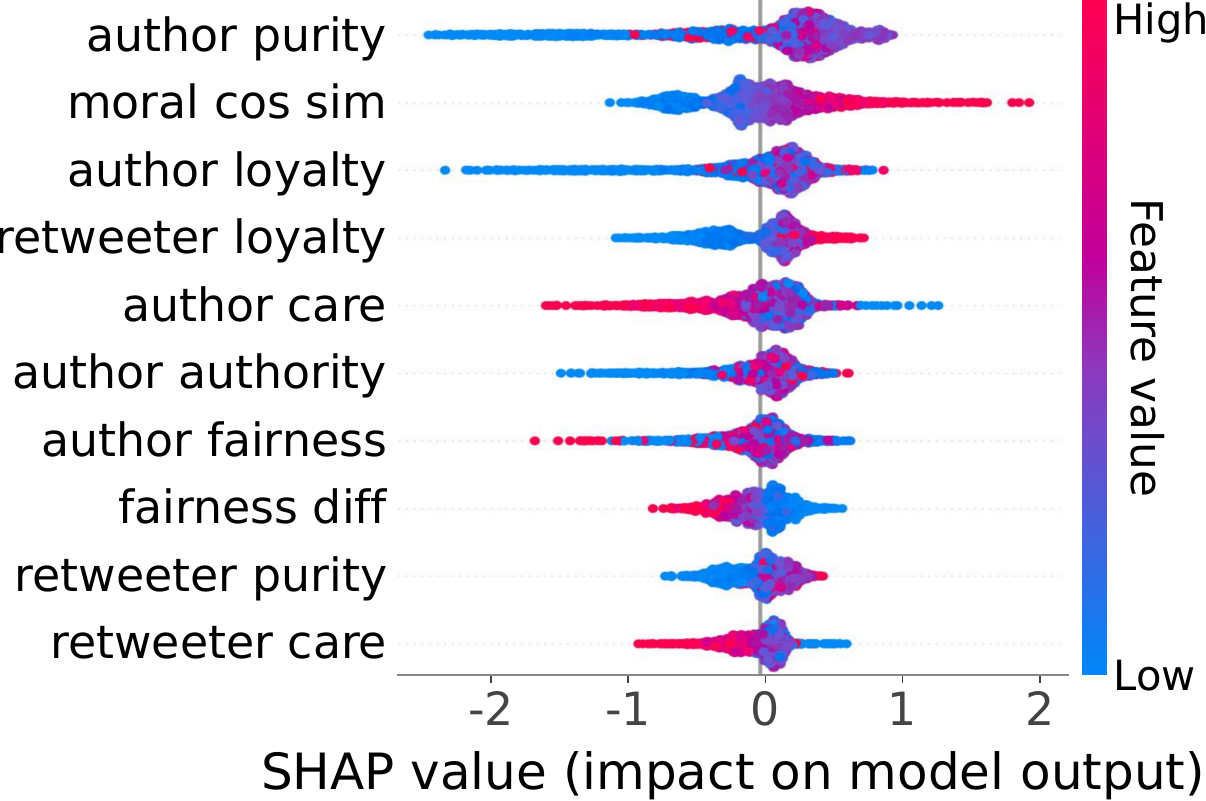}%
    \label{subfig:shap-moral}%
  }%
  \hfill
  \subfloat[]{\label{fig:assortativity}%
    \includegraphics[height=\myheight\textheight]{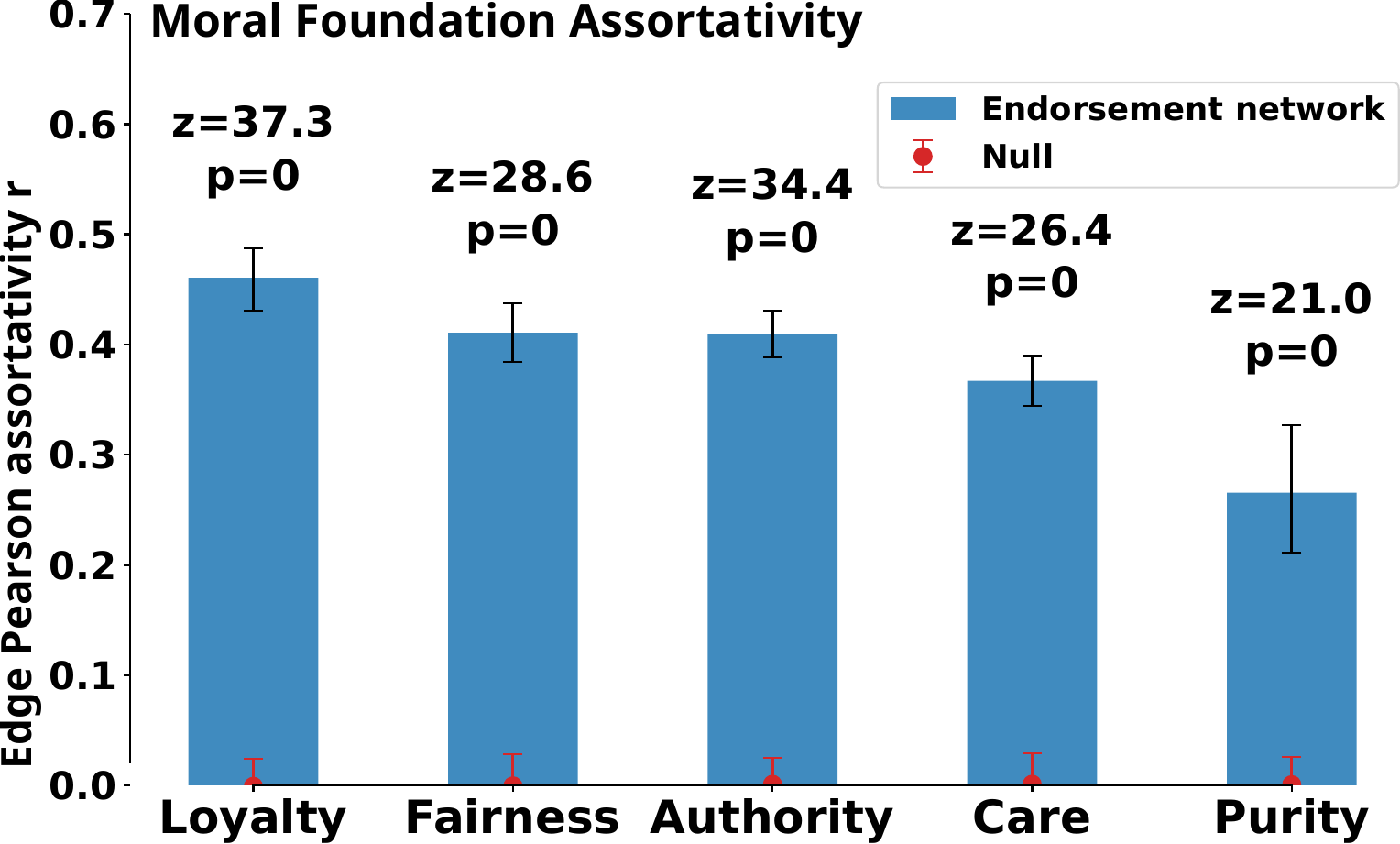}%
    \label{subfig:assortativity}%
  }%
  \caption{Endorsement analysis. (a)~SHAP feature importance for the full prediction model; (b)~SHAP for the moral-only model; (c)~moral assortativity coefficients compared with degree-preserving random networks (edges rewired while preserving each node's in- and out-degree). Error bars in (c) indicate bootstrap confidence intervals.}
  \label{fig:endorsement-shap}
\end{figure*}

\paragraph{Finding 1: Moral content predicts endorsement, with content producer framing carrying stronger signal than consumer-side.}
To evaluate moral alignment in endorsement behavior, we formulate endorsement prediction as a binary classification task over directed user pairs from the retweet network. 
Positive samples correspond to repeated retweet interactions occurring more than five times, while negative samples are constructed using a two-hop non-endorsement constraint.
Endorsed pairs exhibit higher moral cosine similarity than non-endorsed pairs (mean $.934$ vs.\ $.892$, $\Delta=.042$), with both Mann--Whitney and KS tests indicating significant differences ($p<10^{-119}$); the separation is consistent across all five foundations, with loyalty and fairness showing the sharpest distributional gap (\cref{app:homophily-kde}).
We further test whether this structure provides predictive utility beyond network and behavioral signals using a LightGBM classifier with moral-profile, interaction, behavioral, and semantic features. 
More training details are provided in \cref{endorsement-training-details}. As shown in \cref{tab:endorsement-ablation}, removing moral features consistently degrades performance, whereas models using only moral representations still retain substantial predictive power.
Moreover, author-only moral features substantially outperform retweeter-only features, indicating that endorsement behavior is more strongly associated with the moral framing of content producers than with the inferred moral profiles of endorsers.



\paragraph{SHAP analysis.}
In the full model (\cref{subfig:shap-full}), behavioral and network features dominate overall, yet author purity (rank~3) and moral Euclidean distance (rank~5) appear among the top ten, confirming that moral framing adds predictive signal beyond engagement covariates.
In the moral-only model (\cref{subfig:shap-moral}), author-side features hold seven of the top ten positions, with author purity and moral cosine similarity ranking first and second.
Notably, author purity is the strongest individual moral predictor of endorsement despite purity having the lowest network assortativity ($r=.27$), suggesting that high purity framing in content attracts broad endorsement rather than congruent followers.


\paragraph{Finding 2: Endorsement networks exhibit robust moral homophily.}

We next test whether endorsement structure exhibits moral homophily. Following~\citet{PhysRevLett.89.208701}, we measure moral assortativity for each foundation as the edge-level Pearson correlation between the foundation scores of the two endpoints of each endorsement tie. The full computation is provided in \cref{tab:Moral assortativity computation}. 
As shown in \cref{subfig:assortativity}, all five foundations are significantly positively assortative (all $z > 21$, $p \approx 0$ against a degree-preserving null), indicating that users systematically endorse others with similar moral profiles.
Assortativity is highest for loyalty ($r = .46$), followed by fairness and authority ($r = .41$), care ($r = .37$), and purity ($r = .27$).
Per-foundation moral alignment distributions are provided in \cref{app:homophily-kde}.

\paragraph{Human validation of case-study predictions.}

We validate the CHARM-inferred moral score reliability in the COVID-19 domain using 100 stratified tweets independently annotated by three annotators blinded to CHARM's
predictions.
Annotation reliability is high overall, with an observed agreement of .83 and Gwet's AC1 of .76.
Against the majority-voted human labels, CHARM achieves a macro AUC of .84 and a macro average precision of .56.
These results support the use of CHARM-derived moral scores in the case study, although estimates for low-prevalence moral categories should be interpreted cautiously.
Full annotation procedures, agreement statistics, and label-level alignment results are reported in \cref{tab:case-validation} in \cref{app:case-validation}.

\section{Conclusions}
We introduced CHARM, a lightweight moral foundation detection framework that unifies rationale alignment, MAC-based cooperative signals, and hate-speech-aware polarity modeling within a fine-tuned LLM architecture. 
Across multiple datasets, CHARM demonstrates strong robustness, faithfulness, and sample efficiency, while remaining substantially more efficient than prompt-based LLM approaches.
Using CHARM, we further show that online endorsement behavior is closely associated with latent moral alignment, particularly the moral framing of content authors. Notably, author purity is the strongest individual predictor of endorsement yet exhibits the weakest network assortativity ($r=.27$), suggesting that morally charged content can attract broad endorsement beyond moral congruence.
By grounding model components in explicit psychological construct, CHARM bridges social-psychological theory and LLM practice.

\section*{Limitations}
Several limitations should be noted. First, polarity-level moral annotations remain limited, and most existing datasets do not cover the liberty/oppression foundation. As a result, training and evaluation of our 10-dimensional setting were largely restricted to MFTCXplain and HateBRMoralXplain. Future work could improve this by developing larger datasets with polarity-level annotations and broader foundation coverage.

Second, although CHARM shows strong out-of-domain performance on several datasets, results are weaker on VIG and MIC compared to prompt-based approach, whose content focuses more on abstract social norms. In contrast, the in-domain training data mainly consists of concrete, opinion-rich social media and news discourse. Since fine-tuning tends to adapt models toward the training distribution, this mismatch in content type may limit generalization to more abstract forms of moral reasoning.

Finally, MAC supervision relies on \textbf{eMACDscore} automatically generated labels because no publicly available human-annotated MAC dataset currently exists.
Although the same process was applied consistently across all samples, the generated labels may still introduce noise into downstream training.



\section*{Ethics Statement}
This work studies moral reasoning discourse using publicly available datasets collected from social media, news, and online discussion platforms. Since moral annotations can reflect cultural and annotator-specific biases, model predictions should not be interpreted as objective judgments of morality. Our goal is to support the analysis of moral framing and moral conflict in language, rather than to determine whether individuals or viewpoints are morally right or wrong.

The proposed framework may also inherit biases present in the training data, especially in politically or culturally sensitive contexts. We therefore encourage careful use of such models and emphasize that they should not be deployed as standalone systems for high-stakes moderation or decision-making.

\section*{Acknowledgments}

This work was partially supported by the Defence Science and Technology Group (DSTG) and the Advanced Strategic Capabilities Accelerator (ASCA) through its Emerging and Disruptive Technologies Program, and by the Australian Academy of Science.

We thank Lin Tian for her guidance on experiment design and for her contributions to revising the manuscript.

The authors’ contributions are listed below.
\noindent
\textbf{Huixiang Fu:} Conceptual framing, experiment design, dataset sourcing, conducting experiments, result analysis, and writing.

\noindent
\textbf{Marian-Andrei Rizoiu:} Conceptual framing, supervision, funding acquisition, manuscript revision, and writing.

\bibliography{custom}

@article{graham2009liberals,
  title={Liberals and conservatives rely on different sets of moral foundations.},
  author={Graham, Jesse and Haidt, Jonathan and Nosek, Brian A},
  journal={Journal of personality and social psychology},
  volume={96},
  number={5},
  pages={1029},
  year={2009},
  publisher={American Psychological Association},
  doi = {10.1037/a0015141}
}

@article{hoover2020moral,
  author = {
    Joe Hoover and
    Gwenyth Portillo-Wightman and
    Leigh Yeh and
    Shreya Havaldar and
    Aida Mostafazadeh Davani and
    Ying Lin and
    Brendan Kennedy and
    Mohammad Atari and
    Zahra Kamel and
    Madelyn Mendlen
  },
  title = {Moral Foundations Twitter Corpus:
           A Collection of 35k Tweets Annotated for Moral Sentiment},
  journal = {Social Psychological and Personality Science},
  volume = {11},
  number = {8},
  pages = {1057--1071},
  year = {2020}
}

@article{haidt2004intuitive,
  title={Intuitive ethics: How innately prepared intuitions generate culturally variable virtues},
  author={Haidt, Jonathan and Joseph, Craig},
  journal={Daedalus},
  volume={133},
  number={4},
  pages={55--66},
  year={2004},
  publisher={JSTOR}
}

@incollection{graham2013moral,
  title={Moral foundations theory: The pragmatic validity of moral pluralism},
  author={Graham, Jesse and Haidt, Jonathan and Koleva, Sena and Motyl, Matt and Iyer, Ravi and Wojcik, Sean P and Ditto, Peter H},
  booktitle={Advances in experimental social psychology},
  volume={47},
  pages={55--130},
  year={2013},
  publisher={Elsevier}
}

@article{curry2019mapping,
  title={Mapping morality with a compass: Testing the theory of ‘morality-as-cooperation'with a new questionnaire},
  author={Curry, Oliver Scott and Chesters, Matthew Jones and Van Lissa, Caspar J},
  journal={Journal of Research in Personality},
  volume={78},
  pages={106--124},
  year={2019},
  publisher={Elsevier}
}

@article{skorski2025moral,
  title={The Moral Gap of Large Language Models},
  author={Skorski, Maciej and Landowska, Alina},
  journal={arXiv preprint arXiv:2507.18523},
  year={2025}
}

@inproceedings{trager2025mftcxplain,
    title = "{MFTCX}plain: A Multilingual Benchmark Dataset for Evaluating the Moral Reasoning of {LLM}s through Multi-hop Hate Speech Explanation",
    author = "Trager, Jackson  and
      Vargas, Francielle  and
      Alves, Diego  and
      Guida, Matteo  and
      Ngueajio, Mikel K.  and
      Agrawal, Ameeta  and
      Daryani, Yalda  and
      Malekabadi, Farzan Karimi  and
      Plaza-del-Arco, Flor Miriam",
    editor = "Christodoulopoulos, Christos  and
      Chakraborty, Tanmoy  and
      Rose, Carolyn  and
      Peng, Violet",
    booktitle = "Findings of the Association for Computational Linguistics: EMNLP 2025",
    month = nov,
    year = "2025",
    address = "Suzhou, China",
    publisher = "Association for Computational Linguistics",
    url = "https://aclanthology.org/2025.findings-emnlp.851/",
    doi = "10.18653/v1/2025.findings-emnlp.851",
    pages = "15709--15740",
    ISBN = "979-8-89176-335-7",
}

@inproceedings{vargas2026selfexplaining,
    title = "Self-Explaining Hate Speech Detection with Moral Rationales",
    author = "Vargas, Francielle  and
      Trager, Jackson  and
      Alves, Diego  and
      Guida, Matteo  and
      Thapa, Surendrabikram  and
      At{\i}l, Berk  and
      Dementieva, Daryna  and
      Smart, Andrew J  and
      Agrawal, Ameeta",
    editor = "Liakata, Maria  and
      Moreira, Viviane P.  and
      Zhang, Jiajun  and
      Jurgens, David",
    booktitle = "Findings of the {A}ssociation for {C}omputational {L}inguistics: {ACL} 2026",
    month = jul,
    year = "2026",
    address = "San Diego, California, United States",
    publisher = "Association for Computational Linguistics",
    url = "https://aclanthology.org/2026.findings-acl.1704/",
    doi = "10.18653/v1/2026.findings-acl.1704",
    pages = "34109--34131",
    ISBN = "979-8-89176-395-1"
}

@inproceedings{perez2018film,
  title={{FiLM}: Visual Reasoning with a General Conditioning Layer},
  author={Perez, Ethan and Strub, Florian and De Vries, Harm and Dumoulin, Vincent and Courville, Aaron},
  booktitle={Proceedings of the AAAI Conference on Artificial Intelligence},
  year={2018}
}

@article{graham2012mfd,
title={Moral foundations dictionary},
author={Graham, Jesse and Haidt, Jonathan and Nosek, Brian A},
journal={PsycTESTS Dataset},
year={2009},
publisher={American Psychological Association (APA)}
}

@article{trager2022moral,
  author = {
    Jackson Trager and
    Alireza S. Ziabari and
    Aida Mostafazadeh Davani and
    Preni Golazizian and
    Farzan Karimi-Malekabadi and
    Ali Omrani and
    Zhihe Li and
    Brendan Kennedy and
    Nils Karl Reimer and
    Melissa Reyes
  },
  title = {The Moral Foundations Reddit Corpus},
  journal = {arXiv preprint arXiv:2208.05545},
  year = {2022}
}

@inproceedings{deyoung2020eraser,
    title = "{ERASER}: {A} Benchmark to Evaluate Rationalized {NLP} Models",
    author = "DeYoung, Jay  and
      Jain, Sarthak  and
      Rajani, Nazneen Fatema  and
      Lehman, Eric  and
      Xiong, Caiming  and
      Socher, Richard  and
      Wallace, Byron C.",
    editor = "Jurafsky, Dan  and
      Chai, Joyce  and
      Schluter, Natalie  and
      Tetreault, Joel",
    booktitle = "Proceedings of the 58th Annual Meeting of the Association for Computational Linguistics",
    month = jul,
    year = "2020",
    address = "Online",
    publisher = "Association for Computational Linguistics",
    url = "https://aclanthology.org/2020.acl-main.408/",
    doi = "10.18653/v1/2020.acl-main.408",
    pages = "4443--4458",
}

@inproceedings{nguyen2024mformer,
  title={Measuring Moral Dimensions in Social Media with {MFormer}},
  author={Nguyen, Tuan Dung and Chen, Ziyu and Carroll, Nicholas George and Tran, Alasdair and Klein, Colin and Xie, Lexing},
  booktitle={Proceedings of the International AAAI Conference on Web and Social Media},
  volume={18},
  pages={1134--1147},
  year={2024}
}

@inproceedings{chen2025mova,
    title = "{M}o{V}a: Towards Generalizable Classification of Human Morals and Values",
    author = "Chen, Ziyu  and
      Sun, Junfei  and
      Li, Chenxi  and
      Nguyen, Tuan Dung  and
      Yao, Jing  and
      Yi, Xiaoyuan  and
      Xie, Xing  and
      Tan, Chenhao  and
      Xie, Lexing",
    editor = "Christodoulopoulos, Christos  and
      Chakraborty, Tanmoy  and
      Rose, Carolyn  and
      Peng, Violet",
    booktitle = "Proceedings of the 2025 Conference on Empirical Methods in Natural Language Processing",
    month = nov,
    year = "2025",
    address = "Suzhou, China",
    publisher = "Association for Computational Linguistics",
    url = "https://aclanthology.org/2025.emnlp-main.1687/",
    doi = "10.18653/v1/2025.emnlp-main.1687",
    pages = "33216--33260",
    ISBN = "979-8-89176-332-6",
}

@inproceedings{preniqi2024moralbert,
  title={MoralBERT: a fine-tuned language model for capturing moral values in social discussions},
  author={Preniqi, Vjosa and Ghinassi, Iacopo and Ive, Julia and Saitis, Charalampos and Kalimeri, Kyriaki},
  booktitle={Proceedings of the 2024 International Conference on Information Technology for Social Good},
  pages={433--442},
  year={2024}
}

@article{hopp2021extended,
  title={The extended Moral Foundations Dictionary (eMFD): Development and applications of a crowd-sourced approach to extracting moral intuitions from text},
  author={Hopp, Frederic R and Fisher, Jacob T and Cornell, Devin and Huskey, Richard and Weber, Ren{\'e}},
  journal={Behavior research methods},
  volume={53},
  number={1},
  pages={232--246},
  year={2021},
  publisher={Springer}
}

@inproceedings{forbes2020social,
    title = "Social Chemistry 101: Learning to Reason about Social and Moral Norms",
    author = "Forbes, Maxwell  and
      Hwang, Jena D.  and
      Shwartz, Vered  and
      Sap, Maarten  and
      Choi, Yejin",
    editor = "Webber, Bonnie  and
      Cohn, Trevor  and
      He, Yulan  and
      Liu, Yang",
    booktitle = "Proceedings of the 2020 Conference on Empirical Methods in Natural Language Processing (EMNLP)",
    month = nov,
    year = "2020",
    address = "Online",
    publisher = "Association for Computational Linguistics",
    url = "https://aclanthology.org/2020.emnlp-main.48/",
    doi = "10.18653/v1/2020.emnlp-main.48",
    pages = "653--670",
}

@article{clifford2015moral,
  title={Moral foundations vignettes: A standardized stimulus database of scenarios based on moral foundations theory},
  author={Clifford, Scott and Iyengar, Vijeth and Cabeza, Roberto and Sinnott-Armstrong, Walter},
  journal={Behavior research methods},
  volume={47},
  number={4},
  pages={1178--1198},
  year={2015},
  publisher={Springer}
}

@inproceedings{kobbe2020exploring,
  title={Exploring Morality in Argumentation},
  author={Kobbe, Jonathan and Rehbein, Ines and Hulpus, Ioana and Stuckenschmidt, Heiner},
  booktitle={Proceedings of the 7th Workshop on Argument Mining},
  pages={30--40},
  year={2020}
}

@inproceedings{ziems2022moral,
  title={The moral integrity corpus: A benchmark for ethical dialogue systems},
  author={Ziems, Caleb and Yu, Jane and Wang, Yi-Chia and Halevy, Alon and Yang, Diyi},
  booktitle={Proceedings of the 60th Annual Meeting of the Association for Computational Linguistics (Volume 1: Long Papers)},
  pages={3755--3773},
  year={2022}
}

@article{lyu2024faithful,
    title = "Towards Faithful Model Explanation in {NLP}: A Survey",
    author = "Lyu, Qing  and
      Apidianaki, Marianna  and
      Callison-Burch, Chris",
    journal = "Computational Linguistics",
    volume = "50",
    number = "2",
    month = jun,
    year = "2024",
    address = "Cambridge, MA",
    publisher = "MIT Press",
    url = "https://aclanthology.org/2024.cl-2.6/",
    doi = "10.1162/coli_a_00511",
    pages = "657--723",
}

@article{iyer2012understanding,
    doi = {10.1371/journal.pone.0042366},
    author = {Iyer, Ravi AND Koleva, Spassena AND Graham, Jesse AND Ditto, Peter AND Haidt, Jonathan},
    journal = {PLOS ONE},
    publisher = {Public Library of Science},
    title = {Understanding Libertarian Morality: The Psychological Dispositions of Self-Identified Libertarians},
    year = {2012},
    month = {08},
    volume = {7},
    url = {https://doi.org/10.1371/journal.pone.0042366},
    pages = {1-23},
    number = {8},

}

@inproceedings{araque2022libertymfd,
author = {Araque, Oscar and Gatti, Lorenzo and Kalimeri, Kyriaki},
title = {LibertyMFD: A Lexicon to Assess the Moral Foundation of Liberty.},
year = {2022},
isbn = {9781450392846},
publisher = {Association for Computing Machinery},
address = {New York, NY, USA},
url = {https://doi.org/10.1145/3524458.3547264},
doi = {10.1145/3524458.3547264},
booktitle = {Proceedings of the 2022 ACM Conference on Information Technology for Social Good},
pages = {154–160},
numpages = {7},
location = {Limassol, Cyprus},
series = {GoodIT '22}
}

@inproceedings{reinig2024survey,
    title = "A Survey on Modelling Morality for Text Analysis",
    author = "Reinig, Ines  and
      Becker, Maria  and
      Rehbein, Ines  and
      Ponzetto, Simone",
    editor = "Ku, Lun-Wei  and
      Martins, Andre  and
      Srikumar, Vivek",
    booktitle = "Findings of the Association for Computational Linguistics: ACL 2024",
    month = aug,
    year = "2024",
    address = "Bangkok, Thailand",
    publisher = "Association for Computational Linguistics",
    url = "https://aclanthology.org/2024.findings-acl.245/",
    doi = "10.18653/v1/2024.findings-acl.245",
    pages = "4136--4155",
}

@article{malik2025emacd,
author = {Musa Malik and Sungbin Youk and Frederic R. Hopp and Oliver Scott Curry and Marc Cheong and Mark Alfano and René Weber},
title = {The Extended Morality as Cooperation Dictionary (eMACD): A Crowd-Sourced Approach via the Moral Narrative Analyzer Platform},
journal = {Communication Methods and Measures},
volume = {19},
number = {3},
pages = {201--231},
year = {2025},
publisher = {Routledge},
doi = {10.1080/19312458.2025.2500329},
URL = { https://doi.org/10.1080/19312458.2025.2500329},
eprint = {https://doi.org/10.1080/19312458.2025.2500329}
}

@Inbook{curry2016morality,
author="Curry, Oliver Scott",
title="Morality as Cooperation: A Problem-Centred Approach",
bookTitle="The Evolution of Morality",
year="2016",
publisher="Springer International Publishing",
address="Cham",
pages="27--51",
isbn="978-3-319-19671-8",
doi="10.1007/978-3-319-19671-8_2",
url="https://doi.org/10.1007/978-3-319-19671-8_2"
}

@article{haidt2001emotional,
	author = {Jonathan Haidt},
	doi = {10.1037/0033-295x.108.4.814},
	journal = {Psychological Review},
	number = {4},
	pages = {814--834},
	title = {The Emotional Dog and its Rational Tail: A Social Intuitionist Approach to Moral Judgment},
	volume = {108},
	year = {2001}
}

@article{RAO2025,
  author  = {Ashwin Rao and Nazanin Sabri and Siyi Guo and
             Louiqa Raschid and Kristina Lerman},
  title   = {Public Health Messaging on Twitter During the COVID-19 Pandemic:
             Observational Study},
  journal = {Journal of Medical Internet Research},
  volume  = {27},
  year    = {2025}
}

@inproceedings{hendrycks2021aligning,
  title={Aligning {AI} with Shared Human Values},
  author={Hendrycks, Dan and Burns, Collin and Basart, Steven and Critch, Andrew and Li, Jerry and Song, Dawn and Steinhardt, Jacob},
  booktitle={International Conference on Learning Representations (ICLR)},
  year={2021}
}

@article{PhysRevLett.89.208701,
  title = {Assortative Mixing in Networks},
  author = {Newman, M. E. J.},
  journal = {Phys. Rev. Lett.},
  volume = {89},
  issue = {20},
  pages = {208701},
  numpages = {4},
  year = {2002},
  month = {Oct},
  publisher = {American Physical Society},
  doi = {10.1103/PhysRevLett.89.208701},
  url = {https://link.aps.org/doi/10.1103/PhysRevLett.89.208701}
}

@article{Abdurahman2025TargetingAM,
  title={Targeting Audiences' Moral Values Shapes Misinformation Sharing},
  author={Abdurahman, Suhaib and Reimer, Nils K. and Golazizian, Preni and Baek, Elisa and Shen, Yixuan and Trager, Jackson and Lulla, Roshni and Kaplan, Jonas and Parkinson, Carolyn and Dehghani, Morteza},
  journal={Journal of Experimental Psychology: General},
  volume={154},
  number={4},
  pages={935--957},
  year={2025},
  publisher={American Psychological Association},
  doi={10.1037/xge0001714}
}

@inproceedings{amirizaniani2024norm,
  title={{"As an AI Language Model, Yes I Would Recommend Calling the Police"}: Norm Inconsistency in {LLM} Decision-Making},
  author={Jain, Shomik and Calacci, Dan and Wilson, Ashia},
  booktitle={Proceedings of the AAAI/ACM Conference on AI, Ethics, and Society},
  pages={624--633},
  year={2024}
}

@article{jiang2025delph,
  author = {
    Liwei Jiang and
    Jena D. Hwang and
    Chandra Bhagavatula and
    Ronan Le Bras and
    Jenny T. Liang and
    Sydney Levine and
    Jesse Dodge and
    Keisuke Sakaguchi and
    Maxwell Forbes and
    Jack Hessel
  },
  title = {Investigating Machine Moral Judgement through the Delphi Experiment},
  journal = {Nature Machine Intelligence},
  volume = {7},
  number = {1},
  pages = {145--160},
  year = {2025}
}

@misc{frimer2019moral,
  author       = {Jeremy A. Frimer and Reihane Boghrati and Jonathan Haidt and Jesse Graham and Morteza Dehghani},
  title        = {Moral Foundations Dictionary 2.0},
  year         = {2019},
  howpublished = {\url{https://osf.io/ezn37}}
}

@article{kennedy2023moral,
    author = {Kennedy, Brendan and Golazizian, Preni and Trager, Jackson and Atari, Mohammad and Hoover, Joe and Mostafazadeh Davani, Aida and Dehghani, Morteza},
    title = {The (moral) language of hate},
    journal = {PNAS Nexus},
    volume = {2},
    number = {7},
    pages = {pgad210},
    year = {2023},
    month = {07},
    issn = {2752-6542},
    doi = {10.1093/pnasnexus/pgad210},
    url = {https://doi.org/10.1093/pnasnexus/pgad210},
    eprint = {https://academic.oup.com/pnasnexus/article-pdf/2/7/pgad210/51012572/pgad210.pdf},
}

@article{chen2020tracking,
  title={Tracking Social Media Discourse about the {COVID}-19 Pandemic: Development of a Public Coronavirus Twitter Data Set},
  author={Chen, Emily and Lerman, Kristina and Ferrara, Emilio},
  journal={JMIR Public Health and Surveillance},
  volume={6},
  number={2},
  pages={e19273},
  year={2020}
}

@article{guo2023damf, title={A Data Fusion Framework for Multi-Domain Morality Learning}, volume={17}, url={https://ojs.aaai.org/index.php/ICWSM/article/view/22145}, DOI={10.1609/icwsm.v17i1.22145}, number={1}, journal={Proceedings of the International AAAI Conference on Web and Social Media}, author={Guo, Siyi and Mokhberian, Negar and Lerman, Kristina}, year={2023}, month={Jun.}, pages={281–291} }

@inproceedings{zangari2025me2bert,
    title = "{ME}2-{BERT}: Are Events and Emotions what you need for Moral Foundation Prediction?",
    author = "Zangari, Lorenzo  and
      Greco, Candida M.  and
      Picca, Davide  and
      Tagarelli, Andrea",
    editor = "Rambow, Owen  and
      Wanner, Leo  and
      Apidianaki, Marianna  and
      Al-Khalifa, Hend  and
      Eugenio, Barbara Di  and
      Schockaert, Steven",
    booktitle = "Proceedings of the 31st International Conference on Computational Linguistics",
    month = jan,
    year = "2025",
    address = "Abu Dhabi, UAE",
    publisher = "Association for Computational Linguistics",
    url = "https://aclanthology.org/2025.coling-main.638/",
    pages = "9516--9532",
}

@inproceedings{hu2022lora,
  title={{LoRA}: Low-Rank Adaptation of Large Language Models},
  author={Hu, Edward J. and Shen, Yelong and Wallis, Phillip and Allen-Zhu, Zeyuan and Li, Yuanzhi and Wang, Shean and Wang, Lu and Chen, Weizhu},
  booktitle={International Conference on Learning Representations (ICLR)},
  year={2022}
}

@book{haidt2012righteous,
  title={The Righteous Mind: Why Good People Are Divided by Politics and Religion},
  author={Haidt, Jonathan},
  year={2012},
  publisher={Pantheon Books},
  address={New York}
}

@article{graham2011mapping,
  title={Mapping the Moral Domain},
  author={Graham, Jesse and Nosek, Brian A. and Haidt, Jonathan and Iyer, Ravi and Koleva, Spassena and Ditto, Peter H.},
  journal={Journal of Personality and Social Psychology},
  volume={101},
  number={2},
  pages={366--385},
  year={2011},
  publisher={American Psychological Association},
  doi={10.1037/a0021847}
}

@article{dehghani2016purity,
  title={Purity Homophily in Social Networks},
  author={Dehghani, Morteza and Johnson, Kate and Hoover, Joe and Sagi, Eyal and Garten, Justin and Parmar, Niki Jitendra and Vaisey, Stephen and Iliev, Rumen and Graham, Jesse},
  journal={Journal of Experimental Psychology: General},
  volume={145},
  number={3},
  pages={366--375},
  year={2016},
  publisher={American Psychological Association},
  doi={10.1037/xge0000139}
}

@article{curry2019cooperate,
  title={Is It Good to Cooperate? Testing the Theory of Morality-as-Cooperation in 60 Societies},
  author={Curry, Oliver Scott and Mullins, Daniel Austin and Whitehouse, Harvey},
  journal={Current Anthropology},
  volume={60},
  number={1},
  pages={47--69},
  year={2019},
  publisher={University of Chicago Press},
  doi={10.1086/701478}
}

@article{brady2019ideological,
  title={An Ideological Asymmetry in the Diffusion of Moralized Content on Social Media Among Political Leaders},
  author={Brady, William J. and Wills, Julian A. and Burkart, Dominic and Jost, John T. and Van Bavel, Jay J.},
  journal={Journal of Experimental Psychology: General},
  volume={148},
  number={10},
  pages={1802--1813},
  year={2019},
  publisher={American Psychological Association},
  doi={10.1037/xge0000532}
}

@article{vanbavel2021how,
  title={How Social Media Shapes Polarization},
  author={Van Bavel, Jay J. and Rathje, Steve and Harris, Elizabeth and Robertson, Claire and Sternisko, Anni},
  journal={Trends in Cognitive Sciences},
  volume={25},
  number={11},
  pages={913--916},
  year={2021},
  publisher={Elsevier},
  doi={10.1016/j.tics.2021.07.013}
}

@inproceedings{metaxas2015retweets,
  title={What Do Retweets Indicate? Results from User Survey and Meta-Review of Research},
  author={Metaxas, Panagiotis and Mustafaraj, Eni and Wong, Kily and Zeng, Laura and O'Keefe, Megan and Finn, Samantha},
  booktitle={Proceedings of the International AAAI Conference on Web and Social Media},
  pages={658--661},
  year={2015},
  doi={10.1609/icwsm.v9i1.14661}
}

@misc{ccdh2021dozen,
  author       = {{Center for Countering Digital Hate}},
  title        = {The Disinformation Dozen},
  year         = {2021},
  howpublished = {\url{https://counterhate.com/research/the-disinformation-dozen/}},
  note         = {Accessed: 2026-05-24}
}

\clearpage
\appendix

\section{Additional Experimental Details}

\subsection{Experimental Dataset Details}
\label{app:label-distributions}

\cref{tab:in-domain-datasets,tab:mftcxplain-dimensions} report label prevalence for the in-domain training datasets, while \cref{tab:ood-datasets} summarizes label prevalence for the out-of-domain evaluation datasets.

\begin{table*}[t]
  \centering
  \small
  \setlength{\tabcolsep}{4pt}
  \renewcommand{\arraystretch}{1.5}
  \begin{tabular}{lrrrrrrrr}
    \hline
    \textbf{Dataset} & \textbf{Care} & \textbf{Fairness} & \textbf{Loyalty} & \textbf{Authority} & \textbf{Purity} & \textbf{Avg. Tokens} & \textbf{P90. Tokens} & \textbf{Total Size} \\
    \hline
    MFTC        & 40.7\% & 41.8\% & 37.0\% & 41.5\% & 29.7\% & 31.9 & 50  & 13,995 \\
    MFRC        & 35.5\% & 44.2\% & 19.4\% & 32.5\% & 17.4\% & 47.5 & 92  & 7,155 \\
    News        & 26.5\% & 32.4\% & 33.1\% & 35.4\% & 30.4\% & 31.1 & 50  & 12,920 \\
    MFTCXplain  & 37.3\% & 36.0\% & 27.2\% & 23.2\% & 18.6\% & 49.6 & 85  & 3,245 \\
    \hline
    \textbf{Overall} & \textbf{39.9\%} & \textbf{40.3\%} & \textbf{33.4\%} & \textbf{38.3\%} & \textbf{27.0\%} & \textbf{36.2} & \textbf{63} & \textbf{37,315} \\
    \hline
  \end{tabular}
  \caption{Distribution of moral virtue dimensions and input length statistics across in-domain training datasets. The five moral dimensions correspond to virtue--vice axes. Mean token and P90 token denote the average and 90th-percentile token lengths, respectively.}
  \label{tab:in-domain-datasets}
\end{table*}

\begin{table}[t]
  \centering
  \small
  \renewcommand{\arraystretch}{1.5}
  \begin{tabular}{lr}
    \hline
    \textbf{Label} & \textbf{Percentage} \\
    \hline
    Care        & 8.91\%  \\
    Harm        & 29.34\% \\
    Fairness    & 13.50\% \\
    Cheating    & 24.35\% \\
    Loyalty     & 15.04\% \\
    Betrayal    & 13.50\% \\
    Authority   & 10.29\% \\
    Subversion  & 13.22\% \\
    Purity      & 7.95\%  \\
    Degradation & 11.31\% \\
    Hate Speech & 38.92\% \\
    \hline
    \textbf{Total size} & \textbf{3,245} \\
    \hline
  \end{tabular}
  \caption{Label distribution of MFTCXplain across moral polarity dimensions and hate speech annotations.}
  \label{tab:mftcxplain-dimensions}
\end{table}

\begin{table*}[t]
  \centering
  \small
  \setlength{\tabcolsep}{4pt}
  \renewcommand{\arraystretch}{1.5}
  \begin{tabular}{lrrrrrrrr}
    \hline
    \textbf{Dataset} & \textbf{Authority} & \textbf{Care} & \textbf{Fairness} & \textbf{Loyalty} & \textbf{Purity} & \textbf{Avg. Tokens} & \textbf{P90. Tokens} & \textbf{Total Size} \\
    \hline
    VIG    & 14.78\% & 27.83\% & 14.78\% & 13.91\% & 14.78\% & 17.0 & 19  & 132 \\
    SC     & 9.72\%  & 44.44\% & 16.92\% & 17.80\% & 6.79\%  & 12.1 & 17  & 2,9239 \\
    ARG    & 14.37\% & 41.56\% & 16.56\% & 8.12\%  & 19.69\% & 69.6 & 120 & 320 \\
    MIC    & 17.15\% & 51.19\% & 20.72\% & 19.43\% & 10.86\% & 10.2 & 14  & 1,1375 \\
    HateBR & 6.90\%  & 28.97\% & 22.88\% & 4.21\%  & 22.30\% & 23.5 & 47  & 5,040 \\
    \hline
  \end{tabular}
  \caption{Distribution of moral virtue dimensions and input length statistics across out-of-domain evaluation datasets. The five moral dimensions correspond to virtue--vice axes. Avg. token and P90. token denote the average and 90th-percentile token lengths, respectively.}
  \label{tab:ood-datasets}
\end{table*}

\subsection{Details of eMACDscore for MAC Grounding}
\label{app:emacd}

Our MAC grounding signal is derived from \textbf{eMACDscore}, a Python-based moral mining tool introduced by \citet{malik2025emacd} as part of the extended Morality-as-Cooperation Dictionary (eMACD).
Built upon the Morality-as-Cooperation framework
\citep{curry2016morality}, eMACD provides lexicon-based scores across seven cooperation domains: family, group, reciprocity, heroism, deference, fairness, and property. The lexicon itself was constructed through crowd-sourced annotation using the Moral Narrative Analyzer (MoNA) platform.
We use eMACDscore to automatically generate MAC supervision signals for each training instance. Specifically, for every sample and MAC domain, the tool outputs a probability score $\hat{p}_i$ and a sentiment-polarity score $\hat{s}_i$, which are combined into the MAC vector $g \in \mathbb{R}^7$. We adopt eMACDscore because, to the best of our knowledge, there is currently no publicly available human-annotated MAC benchmark dataset.

Importantly, eMACD is developed independently of Moral Foundations Theory (MFT). Rather than being derived from MFT labels or annotations, it is grounded in a distinct theoretical framework and constructed through an independent crowd-sourcing process. As a result, the MAC
supervision provides a complementary moral signal instead of simply repackaging the MFT information already available to the model.
We refer readers to \citet{malik2025emacd} for full details regarding the dictionary construction process, annotation protocol, and validation experiments. 

\subsection{Baseline Implementation Details}
\label{app:baseline-details}

For baseline comparisons, we reproduce each baseline under its original released configuration, including matching the random seed specified in their code, to ensure a fair comparison.

For evaluation, we follow the metrics commonly used in the original studies of each dataset, reporting macro-AUC and macro-F1. 
For F1, per-label decision thresholds are tuned on a shared validation set and applied unchanged to the test set.
We release the trained checkpoint so that all reported scores can be exactly reproduced by running inference on the released model.

\paragraph{\textsc{MFormer}.}
\textsc{MFormer}~\citep{nguyen2024mformer} is a transformer-based moral foundation classification framework built on \textsc{RoBERTa}-base. The model trains five independent binary classifiers, each corresponding to one moral foundation in a one-vs-rest setting. It is trained on a mixture of annotated datasets spanning social media, news, and online discussion domains. During inference, predictions from the five classifiers are combined to produce the final multi-label moral output.

\paragraph{\textsc{MoVa}.}
\textsc{MoVa}~\citep{chen2025mova} is a prompting-based framework for moral value classification using instruction-tuned GPT-4o mini. It formulates moral foundation detection as a multi-label generation task, where a single prompt is used to predict all moral dimensions jointly. The framework does not require task-specific fine-tuning and instead relies on prompt engineering to guide structured moral predictions. In our experiments, we use the official prompt templates released by the authors for reproduction, and treat \textsc{MoVa} as our primary zero-shot baseline for LLM-based moral reasoning.

\paragraph{Qwen-32B-Instruct.}
We include Qwen-32B-Instruct as a larger zero-shot LLM baseline. We use the same moral-label definitions and evaluation protocol as for the other models, without task-specific fine-tuning. This comparison provides an additional reference for the trade-off between model scale and task-specific supervision.

\paragraph{Fine-tuned GPT-4o mini.}
Following \citet{chen2025mova}, we include a supervised fine-tuned variant of GPT-4o mini trained on 30K labeled instances sampled from the \textsc{MFormer} training corpus. The model is fine-tuned to generate structured moral labels directly from input text without rationale supervision or auxiliary moral signals. This baseline serves as a decoder-only LLM adapted for supervised moral classification.
This setup allows us to evaluate whether fine-tuning a LLM improves over prompting-based inference under comparable supervision data.

\paragraph{\textsc{MoralBERT}.}
\textsc{MoralBERT}~\citep{preniqi2024moralbert} is a BERT-based framework for polarity-aware moral classification across ten virtue--vice dimensions. Unlike standard MFT classifiers that predict only foundation presence, \textsc{MoralBERT} explicitly models both positive (virtue) and negative (vice) polarities using a multi-label classification objective. Its explicit polarity modeling provides a relevant comparison for evaluating fine-grained moral representation learning.

\subsection{Evaluation Metric Definitions}
\label{app:metric-definitions}

We report AUC and F1 scores for both hate speech detection and moral classification tasks.

\paragraph{Hate Speech Detection.}
For hate speech classification, we use the standard binary F1 score:
\[
F1 = \frac{2 \cdot \mathrm{Precision} \cdot \mathrm{Recall}}
{\mathrm{Precision} + \mathrm{Recall}},
\]
where precision and recall are computed with respect to the hate speech class.
We additionally report AUC, which measures the area under the receiver operating characteristic (ROC) curve and evaluates the model's ranking quality across all classification thresholds.

\paragraph{Moral Classification.}
Moral classification is formulated as a multi-label prediction task, where each sample may express multiple moral foundations simultaneously. Let \(L\) denote the set of moral labels. For each label \(l \in L\), we compute a binary F1 score by treating the presence or absence of \(l\) as a separate binary classification problem:
\[
F1_l = \frac{2 \cdot \mathrm{Precision}_l \cdot \mathrm{Recall}_l}
{\mathrm{Precision}_l + \mathrm{Recall}_l}.
\]

The reported Macro-F1 is then obtained by averaging over all moral labels:
\[
\mathrm{Macro\text{-}F1}
=
\frac{1}{|L|}
\sum_{l \in L} F1_l.
\]

Similarly, AUC is computed as the average AUC across all moral labels.
Because moral prediction is inherently multi-label and many instances activate only a small subset of moral dimensions, Macro-F1 can sometimes overestimate overall performance. In particular, labels with sparse positive instances or easier decision boundaries may inflate the averaged score despite limited holistic understanding of moral content. We therefore report both Macro-F1 and AUC to provide a more comprehensive assessment of model behaviour.

Unlike AUC, which is threshold-free, F1 depends on a decision threshold. To ensure a fair comparison, we adopt a unified threshold-selection protocol across all models. Using a held-out validation set shared by all methods, we select, for each label  \(l\), the threshold that maximizes the validation (F1\_l), and apply it unchanged to the test set. This isolates differences in model quality from differences in threshold choice.

\paragraph{Rationale Plausibility}

We compare the model-selected rationale tokens $R_k$ against the human-annotated rationale spans $H_k$ for each instance~$k$. We report IoU-F1 and Token-F1:

\[
\mathrm{IoU\mbox{-}F1}
 = \frac{1}{N}\sum_{k=1}^{N} \frac{|R_k \cap H_k|}{|R_k \cup H_k|},
\]

\[
\mathrm{Token\mbox{-}F1}
= \frac{1}{N}\sum_{k=1}^{N} \frac{2|R_k \cap H_k|}{|R_k|+|H_k|}.
\]

We also report AUPRC to evaluate how well token-importance scores rank gold rationale tokens under severe class imbalance.

\paragraph{Rationale Faithfulness}

Following \citet{deyoung2020eraser}, we report \emph{comprehensiveness} and \emph{sufficiency}. Let $p(\cdot)_y$ denote the model probability for the gold label~$y$, and let $e_k$ denote the extracted rationale for instance~$k$.

Comprehensiveness measures the confidence drop after removing rationale tokens:

\[
\mathrm{Comp}
= \frac{1}{N}\sum_{k=1}^{N}\bigl(p(x_k)_y - p(x_k \setminus e_k)_y\bigr).
\]

Sufficiency evaluates whether the rationale alone preserves enough evidence:

\[
\mathrm{Suff}
= \frac{1}{N}\sum_{k=1}^{N}\bigl(p(x_k)_y - p(e_k)_y\bigr).
\]

Higher comprehensiveness indicates that the rationale contributes substantially to the prediction; lower sufficiency indicates that the rationale alone retains most of the predictive signal.

\subsection{Experimental Setup}
\label{Experimental-Setup}
All experiments were conducted on two NVIDIA A40 GPUs using PyTorch 2.10.0 (CUDA 12.8), Hugging Face Transformers 5.0.0, and PEFT 0.18.1.
CHARM is built on LLaMA-3.1-8B and trained in two stages.
In Stage~1, we attach LoRA adapters (rank $16$, $\alpha=32$, dropout $.1$) to all attention and feed-forward projection matrices, and train the moral classifier and token scorer with a binary cross-entropy moral loss. The LoRA adapters introduce .56\% trainable parameters of the full backbone, keeping the model lightweight.
In Stage~2, we freeze the Stage~1 LoRA encoder and train the fusion heads, optimizing a combined objective that sums the moral loss with hate-speech, rationale, and total-variation (TV) smoothness terms, weighted by $\lambda_{\text{hate}}$, $\lambda_{\text{rat}}$, and $\lambda_{\text{TV}}$ (initialized to $.1$, $.2$, and $.05$, respectively; the moral term has a fixed weight of $1.0$). These three $\lambda$ values are learnable via log parameterisation.
Both stages use the AdamW optimizer with a learning rate of $2\times10^{-4}$ and a maximum sequence length of $128$; we adopt these values following common practice without extensive tuning. Stage~1 is trained for $3$ epochs and Stage~2 for $5$ epochs, with a total training time of approximately one hour.

Stage~1 trains on MFTCXplain together with a $30\%$ subsample of the MFRC, MFTC, and News training pools, while Stage~2 trains on MFTCXplain only.
The $30\%$ subsampling uses a fixed seed ($42$) with distribution-aware weighting toward the MFTCXplain label distribution, followed by a $9{:}1$ stratified train/validation split and the same train/test split as \textsc{Mformer}.
The same MFTCXplain train/validation/test partition is used in both stages, so that no test instance is seen during either stage of training.

\subsection{Cross-Validation Robustness}
\label{app:robustness}

To assess the stability of CHARM, we additionally perform
cross-validation on the two smallest datasets, reporting the mean and standard deviation of macro-F1 and AUC across folds (\cref{tab:robustness}). This is an independent robustness check using a different partitioning from the shared-split protocol in \cref{tab:main-results}, and is therefore not directly comparable to the main results.

AUC remains stable across runs (std $\leq .033$), consistent with our observation that threshold-free ranking is robust under variation.
Macro-F1 exhibits larger variance, as expected given the small sample sizes (ARG: $n{=}320$; VIG: $n{=}132$) and F1's sensitivity to threshold selection.

\begin{table}[t]
  \centering
  \small
  \begin{tabular}{lcc}
    \toprule
    \textbf{Dataset} & \textbf{Macro-F1} & \textbf{AUC} \\
    \midrule
    ARG & .489\,$\pm$\,.079 & .864\,$\pm$\,.026 \\
    VIG & .537\,$\pm$\,.048 & .889\,$\pm$\,.033 \\
    \bottomrule
  \end{tabular}
  \caption{Stability of CHARM over 5-fold cross-validation on the
  small-sample datasets (mean\,$\pm$\,std).}
  \label{tab:robustness}
\end{table}

\subsection{Ablation Results on AUC}
\label{app:ablation-auc}
\begin{table*}[t]
  \centering
  \small
  \renewcommand{\arraystretch}{1.5}
  \setlength{\tabcolsep}{4pt}
  \begin{tabular}{lccccccccc}
    \hline
    \textbf{Variant} & \textbf{MFTCXplain$_{10d}$} & \textbf{MFTCXplain$_{5d}$} & \textbf{MFTC} & \textbf{MFRC} & \textbf{News} & \textbf{ARG} & \textbf{SC} & \textbf{MIC} & \textbf{VIG} \\
    \hline
    baseline         & .81 & .77 & .85 & .84 & .75 & .78 & .69 & .68 & .82 \\
    w/o hate speech  & \textbf{.83} & \textbf{.79} & \textbf{.88} & \textbf{.91} & \textbf{.84} & .84 & .75 & .71 & .87 \\
    w/o MAC   & \textbf{.83} & \textbf{.79} & .84 & .84 & .73 & .80 & \textbf{.79} & \textbf{.75} & .82 \\
    w/o rationale    & .82 & .78 & .87 & .90 & .82 & .86 & .77 & .73 & \textbf{.89} \\
    \textsc{CHARM}   & \textbf{.83} & \textbf{.79} & .87 & \textbf{.91} & .83 & \textbf{.87} & .77 & .73 & \textbf{.89} \\
    \hline
  \end{tabular}
  \caption{AUC across in-domain and out-of-domain datasets in ablation studies. Best result per dataset in \textbf{bold}.}
  \label{tab:ablation-auc}
\end{table*}

\cref{tab:ablation-auc} reports AUC-based ablation results across both in-domain and out-of-domain datasets. Overall, removing rationale supervision consistently degrades performance on most datasets, particularly under cross-domain evaluation, suggesting that rationale-guided representations improve robustness beyond token-level alignment. 

The removal of MAC-enhanced representations leads to the largest performance drops on socially grounded datasets such as \textsc{SC} and \textsc{MIC}, indicating that cooperation-oriented moral structure contributes more strongly in open-ended social reasoning settings. In contrast, the hate-speech auxiliary objective primarily improves performance on politically charged or toxicity-related datasets, especially \textsc{MFTC} and \textsc{MFRC}. 

These results suggest that the three components contribute complementary inductive biases: rationale supervision improves interpretability and robustness, MAC representations improve social-generalization capacity, and hate-speech supervision improves sensitivity to morally salient harmful content.

\subsection{Training-Corpus Contribution}
\label{app:corpus-ablation}

We conduct a leave-one-corpus-out analysis to quantify the contribution of MFRC, MFTC, News, and MFTCXplain within CHARM. Each variant uses the same model configuration and evaluation protocol as the full model. This analysis measures the contribution of each corpus within CHARM, but it is not a controlled architecture-only comparison with the baselines.


\begin{table*}[t]
  \centering
  \small
  \renewcommand{\arraystretch}{1.5}
  \setlength{\tabcolsep}{5pt}
  \begin{tabular}{lcccccccccc}
    \hline
    \textbf{Variant} & \textbf{Avg.} & \textbf{MFTC} & \textbf{MFRC} & \textbf{News} & \textbf{ARG} & \textbf{VIG} & \textbf{SC} & \textbf{MIC} & \textbf{MFTCXpl.} & \textbf{HateBR} \\
    \hline
    \textsc{CHARM} & \textbf{.83} & .87 & \textbf{.91} & .83 & \textbf{.87} & \textbf{.89} & \textbf{.77} & \textbf{.73} & .79 & \textbf{.82} \\
    w/o MFRC & .63 & .66 & .55 & .64 & .59 & .53 & .56 & .54 & .84 & .78 \\
    w/o MFTC & .70 & .55 & .78 & .67 & .73 & .61 & .69 & .61 & .85 & .80 \\
    w/o News & .71 & .70 & .76 & .52 & .72 & .69 & .68 & .62 & \textbf{.86} & .81 \\
    w/o MFTCXplain & .80 & \textbf{.90} & .90 & \textbf{.84} & .82 & \textbf{.89} & .74 & .70 & .70 & .69 \\
    \hline
  \end{tabular}
  \caption{AUC results for the leave-one-corpus-out analysis. The average is computed across all nine evaluation datasets. Best results per dataset are shown in \textbf{bold}.}
  \label{tab:corpus-ablation-auc}
\end{table*}

\begin{table*}[t]
  \centering
  \small
  \renewcommand{\arraystretch}{1.5}
  \setlength{\tabcolsep}{5pt}
  \begin{tabular}{lcccccccccc}
    \hline
    \textbf{Variant} & \textbf{Avg.} & \textbf{MFTC} & \textbf{MFRC} & \textbf{News} & \textbf{ARG} & \textbf{VIG} & \textbf{SC} & \textbf{MIC} & \textbf{MFTCXpl.} & \textbf{HateBR} \\
    \hline
    \textsc{CHARM} & \textbf{.57} & .72 & \textbf{.72} & .55 & \textbf{.54} & \textbf{.67} & \textbf{.46} & \textbf{.50} & \textbf{.59} & \textbf{.38} \\
    w/o MFRC & .42 & .58 & .38 & .49 & .37 & .34 & .33 & .40 & .53 & .37 \\
    w/o MFTC & .49 & .51 & .61 & .53 & .51 & .44 & .44 & .43 & .53 & \textbf{.38} \\
    w/o News & .51 & .65 & .61 & .43 & .52 & .57 & .43 & .45 & .58 & \textbf{.38} \\
    w/o MFTCXplain & .54 & \textbf{.77} & .67 & \textbf{.61} & \textbf{.54} & .59 & .43 & .45 & .47 & .34 \\
    \hline
  \end{tabular}
  \caption{F1 results for the leave-one-corpus-out analysis. The average is computed across all nine evaluation datasets. Best results per dataset are shown in \textbf{bold}.}
  \label{tab:corpus-ablation-f1}
\end{table*}

As shown in \cref{tab:corpus-ablation-auc,tab:corpus-ablation-f1}, the full configuration achieves the strongest average performance. MFRC makes the largest average contribution, while MFTC and News provide complementary domain coverage. Removing MFTCXplain lowers average performance and produces the clearest declines on MFTCXplain and HateBR, confirming the value of its polarity, rationale, and hate-speech supervision. Variation across individual datasets suggests corpus-specific interactions rather than uniform gains from every training source. Overall, the results reflect the complementary value of corpus diversity and richer supervision.

\subsection{Qualitative Rationale Analysis}
\label{app:qualitative-rationales}

We select cases in which CHARM exactly matches the reference foundation set, whereas the model without rationale supervision misses at least one reference foundation. The examples are therefore intended to clarify how rationale supervision affects evidence use rather than provide an additional performance evaluation. For visualization, subword rationale probabilities are averaged within each reconstructed word,
\[
p(w)=\frac{1}{|T(w)|}\sum_{t\in T(w)}p_t,
\]
where $T(w)$ is the set of subword tokens associated with word $w$. A threshold of $p(w)>.5$ is used only for visualization and does not affect moral-label prediction.

\begin{table*}[t]
  \centering
  \small
  \renewcommand{\arraystretch}{1.5}
  \renewcommand{\tabularxcolumn}[1]{m{#1}}
  \setlength{\tabcolsep}{4pt}
  \begin{tabularx}{\textwidth}{>{\centering\arraybackslash}m{0.06\textwidth}>{\raggedright\arraybackslash}X>{\centering\arraybackslash}m{0.18\textwidth}>{\centering\arraybackslash}m{0.18\textwidth}}
    \hline
    \textbf{Case} & \multicolumn{1}{c}{\textbf{Text}} & \textbf{CHARM} & \textbf{w/o rationale} \\
    \hline
    1 & Minor or major, it's all teamwork. You don't want to be part of her team. It's everyone pitching in with whatever, whenever, just because it makes everyone's lives a bit better. & \shortstack[c]{Care/Harm\\Fairness/Cheating\\Loyalty/Betrayal} & \shortstack[c]{Care/Harm\\Fairness/Cheating\\Authority/Subversion} \\
    \hline
    2 & Director James Comey was critical of Clinton's use of the server but said he would not recommend pursuing criminal charges. & Authority/Subversion & Fairness/Cheating \\
    \hline
    3 & The Katana is so sharp, they can simply cut the coronavirus if it infects someone, without any damage whatsoever to the infectee. & \shortstack[c]{Fairness/Cheating\\Purity/Degradation} & Fairness/Cheating \\
    \hline
    \noalign{\vskip 4pt}
    4 & Bloomberg is another stupid Liberal politician; Sandy victims deserve attention and resources, not a marathon. & \shortstack[c]{Care/Harm\\Fairness/Cheating\\Loyalty/Betrayal\\Authority/Subversion} & \shortstack[c]{Fairness/Cheating\\Loyalty/Betrayal\\Authority/Subversion} \\[4pt]
    \hline
  \end{tabularx}
  \caption{Qualitative comparison of CHARM and the model without rationale supervision. CHARM matches the reference foundation set in each example, whereas the ablated model misses at least one reference foundation.}
  \label{tab:rationale-examples}
\end{table*}

As illustrated in \cref{tab:rationale-examples}, CHARM often retains evidence related to suffering, assistance, group loyalty, authority, and contamination. In Case~1, repeated references to ``teamwork,'' ``her team,'' and ``everyone pitching in'' make the Loyalty/Betrayal signal explicit, whereas the ablated model predicts Authority/Subversion. In Case~4, ``Sandy victims'' and their need for ``attention and resources'' emphasize suffering and assistance, helping CHARM recover Care/Harm. These examples complement the quantitative rationale ablation by showing how rationale supervision can preserve label-relevant evidence.

\begin{figure}[t]
  \centering
  \includegraphics[width=0.90\columnwidth]{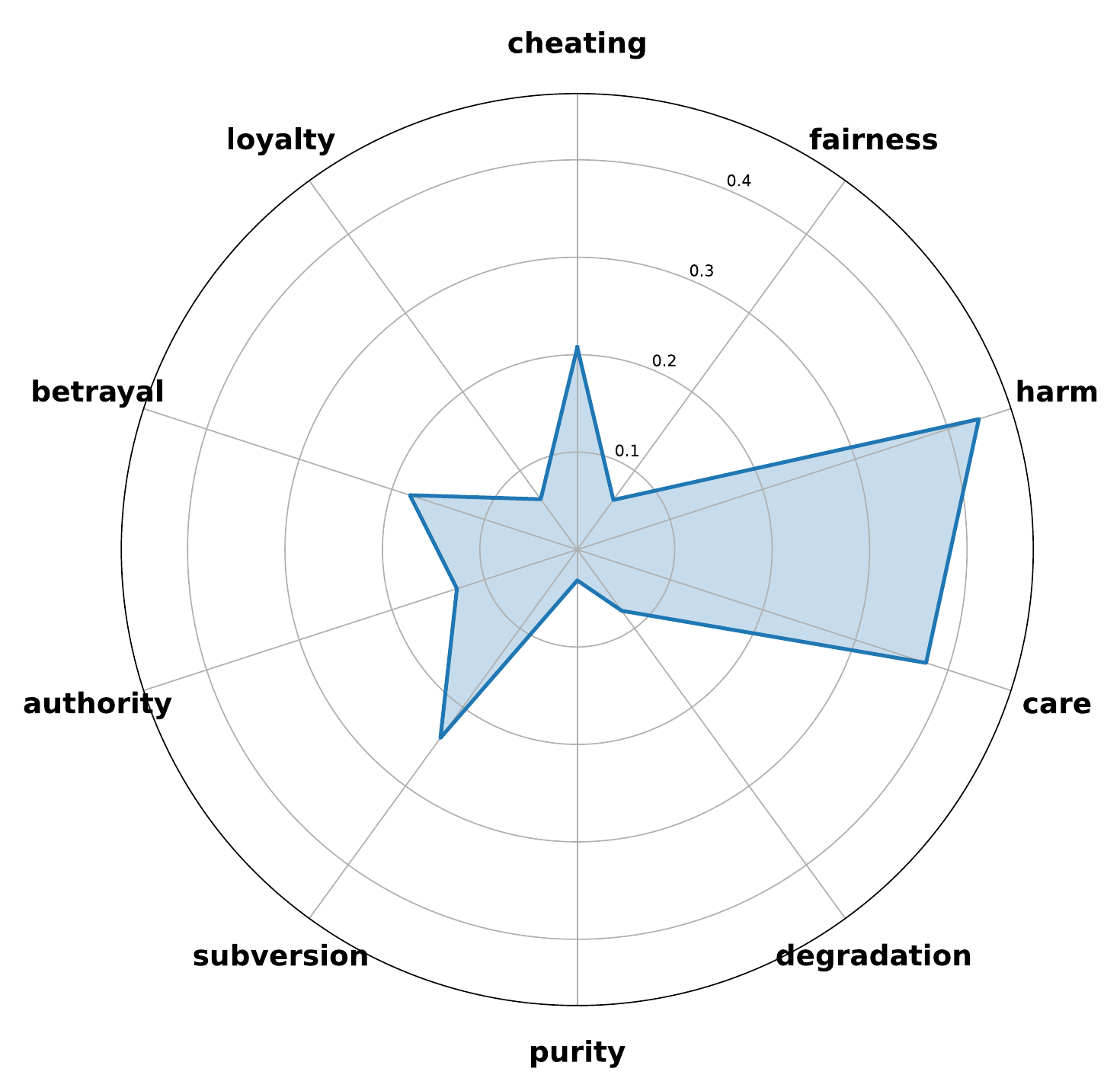}
  \caption{Distribution of users across the ten polarity-aware moral dimensions.}
  \label{fig:radar_10d_positive_nodes_mean_profile}
\end{figure}

\subsection{Per-Foundation Performance Analysis}
\label{app:per-label}

\begin{figure*}[t]
  \centering
  \includegraphics[width=\textwidth]{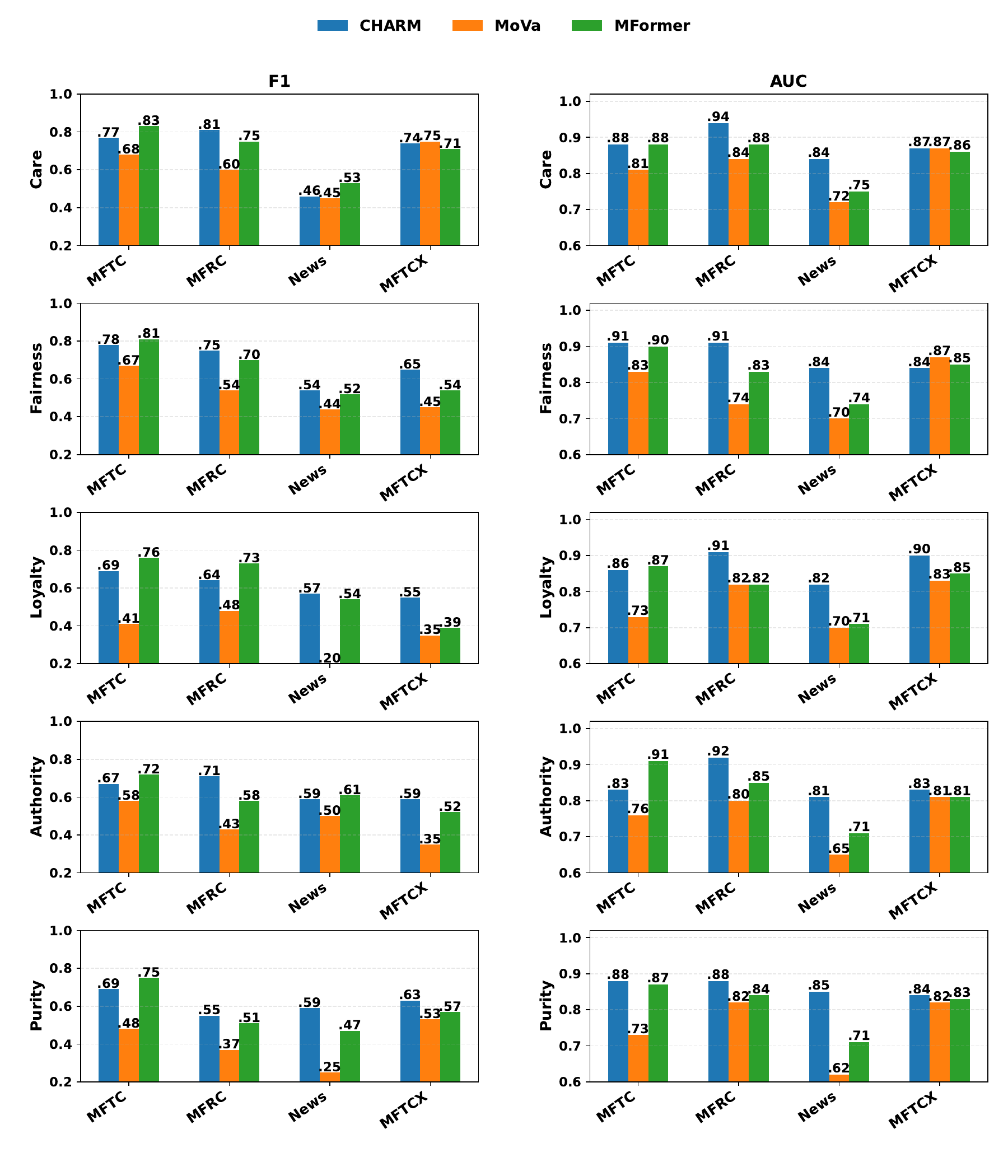}
  \caption{In-domain evaluation: F1 and AUC across datasets.}
  \label{fig:in-domain-f1-auc}
\end{figure*}

\begin{figure*}[t]
  \centering
  \includegraphics[width=\textwidth]{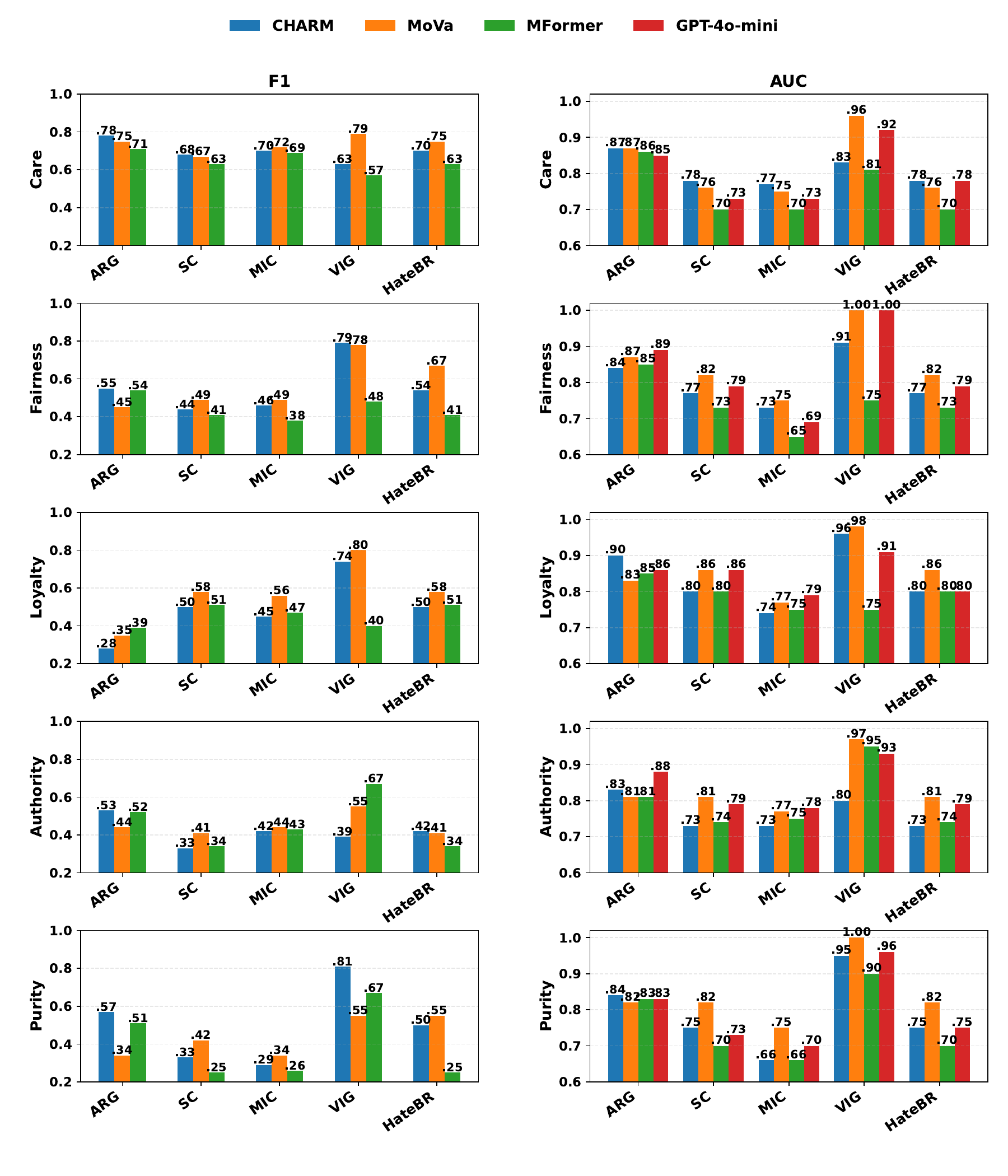}
  \caption{Out-of-domain evaluation: F1 and AUC across datasets.}
  \label{fig:ood-f1-auc}
\end{figure*}




\cref{fig:in-domain-f1-auc} presents dataset-level Macro-F1 and \cref{fig:ood-f1-auc} AUC comparisons under both in-domain and out-of-domain settings. Across datasets, \textsc{CHARM} maintains relatively stable degradation under distribution shift, with larger robustness gaps emerging primarily on short-text and socially ambiguous datasets such as \textsc{SC} and \textsc{MIC}. 

Compared with in-domain evaluation, out-of-domain performance decreases are generally moderate, suggesting that the model captures transferable moral representations rather than relying solely on dataset-specific lexical cues. Notably, AUC scores remain comparatively stable even when F1 declines, indicating that the model preserves ranking consistency under domain shift despite threshold-sensitive classification degradation.

\section{Moral Alignment and Endorsement Prediction}
\label{app:endorsement-details}

\subsection{Endorsement Network Construction}
\label{app:Endorsement-Network-Construction}
We construct a directed endorsement graph from the COVID-19 Twitter dataset introduced by \citet{chen2020tracking}. Retweets are treated as endorsement signals following prior work  \citep{metaxas2015retweets}, while mentions and quote tweets are excluded because they may reflect disagreement or contextual commentary.

For endorsement prediction, positive samples correspond to directed retweet interactions occurring more than five times between two users. Negative samples are constructed using a two-hop non-endorsement constraint to avoid trivially disconnected user pairs. The construction process is illustrated in \cref{fig:endorsement-structure}.

\cref{fig:radar_10d_positive_nodes_mean_profile} shows the distribution of users across the ten polarity-aware moral dimensions. Positive moral poles are substantially denser and more stable than negative poles, while several vice dimensions remain highly sparse in large-scale Twitter discourse. To reduce sparsity and improve robustness, we therefore collapse virtue and vice polarity scores into five foundation-level moral representations when constructing user-level moral profiles for endorsement analysis.

\subsection{Manual Validation of the COVID-19 Case Study}
\label{app:case-validation}

To assess the validity of the moral measurements underlying the downstream
case study, we evaluate CHARM's predictions on a stratified sample of 100
COVID-19 tweets.
Tweets are stratified by predicted moral foundation and moral intensity
to ensure coverage of different predicted moral profiles.
Three annotators independently assign polarity-level moral labels while
blinded to CHARM's predictions.
A label is retained in the human reference when selected by at least two
annotators.
Positive labels account for 17.6\% of all tweet--label decisions.
Given this class imbalance, we report observed agreement together with
Fleiss' $\kappa$, prevalence-adjusted bias-adjusted kappa (PABAK), and
Gwet's AC1.
We additionally evaluate the alignment between CHARM's continuous
predictions and the majority-voted human reference using AUC and average
precision (AP).

\begin{table*}[t]
  \centering
  \small
  \setlength{\tabcolsep}{5pt}
  \begin{tabular}{lrrrrrr}
    \toprule
    \textbf{Label} & \textbf{Pos.} & \textbf{Agreement} & \textbf{Fleiss' $\kappa$} & \textbf{PABAK} & \textbf{AUC} & \textbf{AP} \\
    \midrule
    Care        & 29 & .79 & .49 & .57 & .86 & .73 \\
    Harm        & 37 & .64 & .25 & .28 & .71 & .64 \\
    Fairness    &  2 & .87 & .03 & .75 & .95 & .23 \\
    Cheating    & 20 & .79 & .30 & .57 & .90 & .69 \\
    Loyalty     &  5 & .89 & .35 & .79 & .87 & .44 \\
    Betrayal    &  6 & .85 & .19 & .71 & .71 & .14 \\
    Authority   & 11 & .81 & .27 & .63 & .77 & .49 \\
    Subversion  & 27 & .76 & .43 & .52 & .83 & .72 \\
    Purity      &  5 & .98 & .80 & .96 & 1.00 & 1.00 \\
    Degradation & 12 & .91 & .58 & .83 & .83 & .48 \\
    \midrule
    Overall  & -- & .83 & .42 & .66 & .84 & .56 \\
    \bottomrule
  \end{tabular}
  \caption{Annotation reliability and alignment between CHARM predictions and
  majority-voted human labels on 100 COVID-19 tweets. \textit{Pos.} denotes the number of majority-voted positive examples, and AP denotes average precision. Overall Gwet's AC1 is .760.}
  \label{tab:case-validation}
\end{table*}

\paragraph{Annotation reliability.}

Across all tweet--label decisions, observed agreement is .830, while
Fleiss' $\kappa$ is .415, PABAK is .660, and Gwet's AC1 is .760.
The discrepancy between raw agreement and Fleiss' $\kappa$ is most
pronounced for low-prevalence labels.
For example, only two tweets receive a majority-voted Fairness label,
such that its high observed agreement primarily reflects agreement on
negative cases.
The prevalence-adjusted measures therefore provide complementary evidence
of annotation reliability under the sparse multilabel setting.

\paragraph{Human--model alignment.}

Against the majority-voted human reference, CHARM achieves a polarity-level
macro AUC of .842 with a 95\% bootstrap confidence interval of
[.801, .876], together with a macro average precision of .555.
As shown in \cref{tab:case-validation}, performance varies across moral
categories.
The results indicate that CHARM's continuous predictions generally align
with human judgments in the COVID-19 domain, providing additional support
for their use in the downstream endorsement analyses.

\paragraph{Limitations of label-level estimates.}

Label-level results for low-prevalence categories should be interpreted
cautiously.
In particular, Fairness, Loyalty, Purity, and Betrayal contain only
2, 5, 5, and 6 majority-voted positive examples, respectively.
Consequently, high AUC or agreement values for these categories can be
unstable and may be driven partly by the large number of negative
instances.
We therefore emphasize the aggregate validation results rather than
drawing strong conclusions from individual rare-category estimates.

\subsection{Backbone Comparison Results}
\label{app:qwen-results}
\cref{tab:qwen_charm} compares CHARM with Qwen3-8B across both in-domain and out-of-domain datasets. CHARM achieves consistently stronger performance on most datasets, with particularly large gains on MFTC, MFRC, and News. The improvements are also stable under out-of-domain evaluation, especially on ARG and SC, suggesting that the combination of rationale supervision and MAC grounding provides more robust moral representations than direct prompting alone. While Qwen3-8B remains competitive on VIG, overall results indicate that CHARM offers a stronger balance between efficiency and predictive performance.

\begin{table}[t]
\centering
\small
\renewcommand{\arraystretch}{1.15}
\setlength{\tabcolsep}{4pt}
\begin{tabular}{lcccccc}
\hline
\multicolumn{1}{c}{\textbf{Dataset}} & \multicolumn{2}{c}{\textbf{Qwen3-8B}} & \multicolumn{2}{c}{\textbf{CHARM}} \\[2pt]
\cline{2-5}
 & \textbf{F1} & \textbf{AUC} & \textbf{F1} & \textbf{AUC} \\[2pt]
\hline
MFTC   & .57 & .76 & .72 & .87 \\
MFRC   & .46 & .76 & .72 & .91 \\
News   & .39 & .66 & .55 & .83 \\
ARG    & .48 & .79 & .54 & .87 \\
SC     & .38 & .70 & .46 & .77 \\
MIC    & .40 & .67 & .50 & .73 \\
VIG    & .63 & .90 & .67 & .89 \\
\hline
\end{tabular}
\caption{Performance comparison between Qwen3-8B and CHARM across datasets in terms of Macro-F1 and AUC.}
\label{tab:qwen_charm}
\end{table}

\subsection{User Moral Profile Construction}

User-level moral profiles are constructed by aggregating tweet-level moral predictions inferred by \textsc{CHARM} over the five MFT foundations. Virtue and vice polarity scores are collapsed into unified foundation-level moral intensity scores.
For each user $u$, moral scores are aggregated over high-intensity moral tweets:
\[
m_u^{(k)} = \frac{1}{|T_u^{(k)}|} \sum_{t \in T_u^{(k)}} s_t^{(k)},
\]
where $s_t^{(k)}$ denotes the predicted moral intensity for foundation $k$, and $T_u^{(k)}$ contains tweets whose intensity exceeds the user-specific 90th percentile threshold.
\cref{fig:radar_10d_positive_nodes_mean_profile} visualizes the distribution of users across the ten polarity-aware moral dimensions.

\subsection{Feature Construction}
\label{Feature-Construction}
\cref{tab:feature-list} summarizes the features used for endorsement prediction.

Pairwise moral similarity is computed using cosine similarity:
\[
\mathrm{cos\_sim}(u,v)
=
\frac{\mathbf{m}_u^\top \mathbf{m}_v}
{\|\mathbf{m}_u\|\|\mathbf{m}_v\|},
\]
and Euclidean distance:
\[
\mathrm{euclidean\_dist}(u,v)
=
\|\mathbf{m}_u - \mathbf{m}_v\|_2.
\]

Per-foundation moral differences are defined as:
\[
\mathrm{diff}^{(k)}(u,v)
=
m_u^{(k)} - m_v^{(k)}.
\]

Behavioral statistics are aggregated using:
\[
\mathrm{activity\_mean}(u)
=
\frac{1}{|T_u|}
\sum_{t \in T_u}
a_t,
\]
where $a_t$ denotes tweet-level engagement statistics such as replies, favorites, quotes, and retweets.

\begin{table*}[t]
\centering
\small
\setlength{\tabcolsep}{3pt}
\renewcommand{\arraystretch}{1.1}
\begin{tabularx}{\textwidth}{>{\raggedright\arraybackslash}p{0.16\textwidth} >{\raggedright\arraybackslash}p{0.12\textwidth} >{\raggedright\arraybackslash}p{0.20\textwidth} >{\raggedright\arraybackslash}X}
\toprule
\textbf{Feature Category} & \textbf{Level} & \textbf{Feature} & \textbf{Description} \\
\midrule

\textbf{Moral Features}
& Author / Retweeter
& Care, Fairness, Loyalty, Authority, Purity 
& User-level moral intensity scores aggregated over the five moral foundations using polarity-collapsed representations. \\

\midrule

\multirow{3}{0.16\textwidth}{\textbf{Moral Interaction Features}}
& User Pair
& Moral Similarity 
& Cosine similarity between author and retweeter moral representations. \\

& User Pair
& Moral Distance 
& Euclidean distance between author and retweeter moral representations. \\

& User Pair
& Foundation-Level Moral Difference 
& Per-foundation differences between retweeter and author moral scores. \\

\midrule

\multirow{3}{0.16\textwidth}{\textbf{Behavioral Features}}
& User-Level
& Social Connectivity 
& User engagement statistics including follower count, friend count, and favourites count. \\

& User-Level
& Interaction Activity 
& Average frequencies of replies, quotes, and retweets across authored tweets. \\

& User Pair
& Relative Influence Gap 
& Differences in user influence and activity between author and retweeter. \\

\midrule

\textbf{Semantic Features}
& User Pair
& Biography Similarity 
& Cosine similarity and Euclidean distance between user biography embeddings. \\

\bottomrule
\end{tabularx}

\caption{
Feature definitions for endorsement prediction.
}
\label{tab:feature-list}

\end{table*}

\subsection{Endorsement Prediction Model Training Details}
\label{endorsement-training-details}
We train a binary LightGBM classifier using moral-profile, interaction, behavioral, and semantic features. The model is trained with a binary logistic objective and evaluated across five random seeds.

We use a learning rate of .05, maximum tree depth of 6, 200 boosting rounds, and 64 leaves per tree. Early stopping is applied based on validation-set AUC with a patience of 20 rounds. To reduce overfitting, we apply both feature and row subsampling with rates of .8 together with L2 regularization coefficient 1.0.

All continuous features are z-score normalized before training. Semantic similarity features are computed using cosine similarity between user-level sentence embeddings, while moral similarity features are computed using cosine similarity between aggregated user moral profiles inferred by \textsc{CHARM}. Reported results are averaged across all runs.

\subsection{Per-Foundation Moral Alignment Distributions}
\label{app:homophily-kde}

\begin{figure*}[t]
  \centering
  \includegraphics[width=\textwidth]{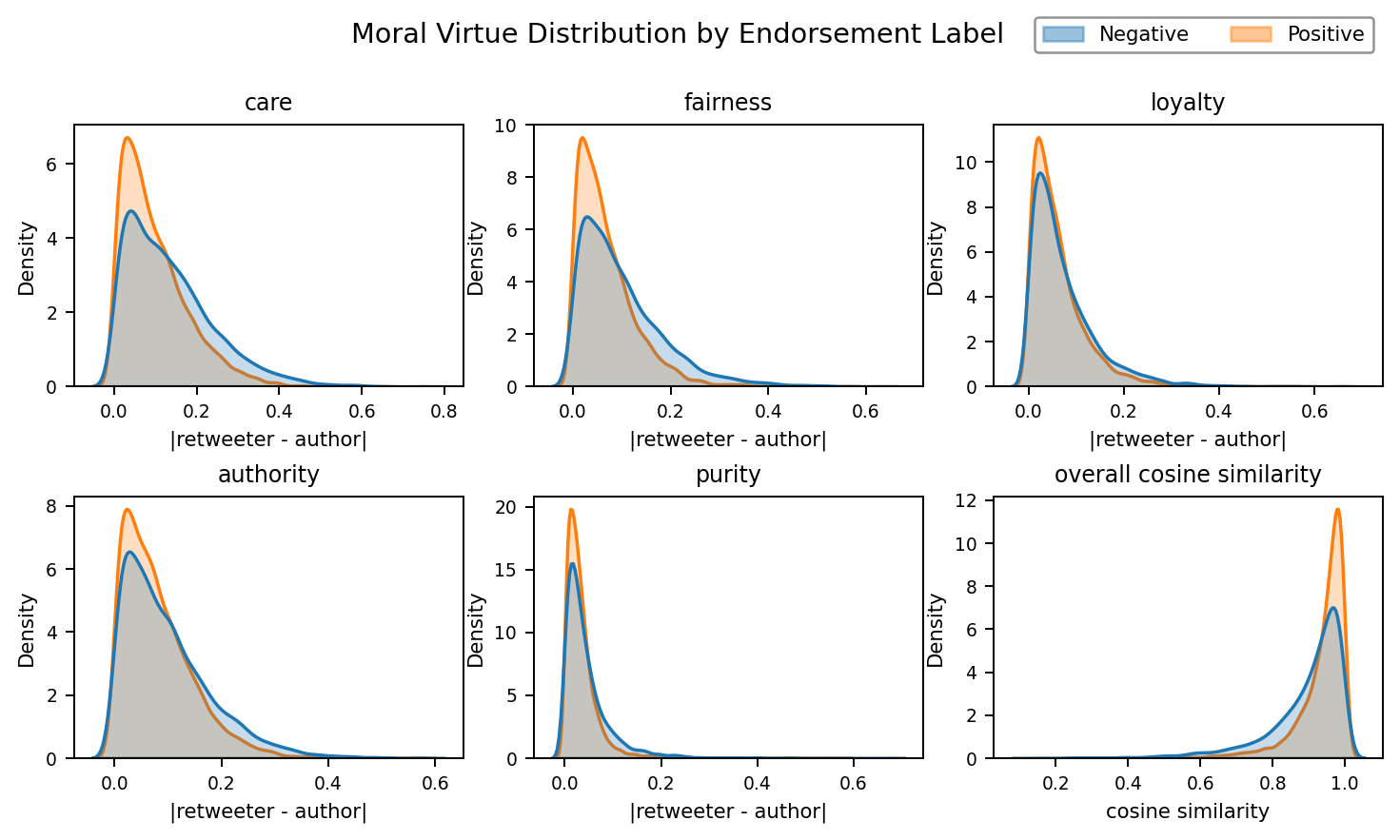}
  \caption{Per-foundation KDE distributions of $|\text{retweeter moral score} - \text{author moral score}|$ for endorsed (positive) and non-endorsed (negative) pairs, across the five MFT foundations and overall cosine similarity (bottom right). Endorsed pairs consistently exhibit smaller per-foundation divergence and higher cosine similarity.}
  \label{fig:homophily-kde}
\end{figure*}

\cref{fig:homophily-kde} shows per-foundation distributions of moral divergence ($|\text{retweeter} - \text{author}|$) and overall cosine similarity for endorsed and non-endorsed pairs.
Across all five foundations, endorsed pairs exhibit tighter moral alignment --- smaller author--retweeter divergence --- than non-endorsed pairs, confirming that homophily holds at the individual-foundation level.
The separation is most pronounced for loyalty and fairness, consistent with the assortativity coefficients in \cref{subfig:assortativity}.

\subsection{Moral Assortativity Computation}
\label{tab:Moral assortativity computation}

We measure moral homophily using Newman assortativity for continuous attributes. Given a directed network with edges $(i \to j) \in E$, each edge is associated with source and target moral scores for a specific foundation.
The assortativity coefficient is computed as the Pearson correlation across edges:
\[
r \;=\;
\frac{\sum_{(i,j)\in E}(x_i-\bar{x}_s)(y_j-\bar{y}_t)}
{\sqrt{\sum_{(i,j)\in E}(x_i-\bar{x}_s)^2}
\sqrt{\sum_{(i,j)\in E}(y_j-\bar{y}_t)^2}}
\]
where $\bar{x}_s$ and $\bar{y}_t$ denote the edge-weighted means of source and target attributes.
We compute assortativity separately for each moral foundation. Statistical significance is assessed using a degree-preserving null model in which edges are randomly rewired while preserving each node's in-degree and out-degree distributions.

\end{document}